\documentclass[runningheads]{llncs}
\usepackage{graphicx}
\usepackage{amssymb}
\usepackage{amsmath}
\usepackage{bm}
\usepackage[small]{caption}
\usepackage[english]{babel}
\usepackage{stmaryrd}
\usepackage[whole]{bxcjkjatype}
\usepackage{times}
\usepackage{algorithm}
\usepackage{algpseudocode}

\DeclareMathOperator*{\argmax}{arg\,max}

\begin{document}
\title{A Bayesian Approach to Direct and Inverse\\
Abstract Argumentation Problems\thanks{This paper was submitted to the journal of Artificial Intelligence (AIJ) and rejected.}}
%
%
\author{Hiroyuki Kido\inst{1} \and
Beishui Liao\inst{2}}
\institute{Sun Yat-sen University, China\\
\email{kido@mail.sysu.edu.cn} \and
Zhejiang University, China\\
\email{baiseliao@zju.edu.cn}}
\begin{sloppypar}
\maketitle
\begin{abstract}
This paper studies a fundamental mechanism of how to detect a conflict between arguments given sentiments regarding acceptability of the arguments. We introduce a concept of the inverse problem of the abstract argumentation to tackle the problem. Given noisy sets of acceptable arguments, it aims to find attack relations explaining the sets well in terms of acceptability semantics. It is the inverse of the direct problem corresponding to the traditional problem of the abstract argumentation that focuses on finding sets of acceptable arguments in terms of the semantics given an attack relation between the arguments. We give a probabilistic model handling both of the problems in a way that is faithful to the acceptability semantics. From a theoretical point of view, we show that a solution to both the direct and inverse problems is a special case of the probabilistic inference on the model. We discuss that the model provides a natural extension of the semantics to cope with uncertain attack relations distributed probabilistically. From en empirical point of view, we argue that it reasonably predicts individuals sentiments regarding acceptability of arguments. This paper contributes to lay the foundation for making acceptability semantics data-driven and to provide a way to tackle the knowledge acquisition bottleneck.
\end{abstract}


\section{Introduction}
The world is full of difficult problems. Argumentation is a human cognitive process to understand them. The driving force of argumentation is a conflict of opinions. The ability to detect conflicts is essential for humans to engage in argumentation. The way humans recognize conflicts between arguments is based on a simple principle of mutual exclusivity or incompatibility regarding acceptability of arguments. Let us take a look at an example to illustrate the principle.
%
\begin{example}
A professor and a government official argue about a government's policy on the allocation of a research budget.
\par
\noindent
\begin{description}
\item[Professor:] The government should widely and fairly allocate research funds for research diversity.
\item[Official:] Our government should select and concentrate on promising research in terms of cost effectiveness.
\end{description}
%
At this point, no explicit conflict is stated in these arguments. The existence of a conflict thus depends on the context of the arguments and knowledge of the arguers or listeners.
\par
Suppose that a rational agent judges these arguments not to be acceptable at the same time no matter how the agent considers acceptability of the individual arguments. This incompatible acceptability of the arguments must lead to agent's opinion that there is a conflict between the arguments.
\end{example}
The important insight we can obtain from this example is that agent's sentiment on the acceptability of the individual arguments is the cause of agent's recognition of a conflict between the arguments. The goal of this paper is to give a scientific account of a detection of a conflict relation between arguments based on the mutual exclusivity of acceptability of arguments.
\par
%
There are at least two computational methodologies to detect a conflict relation between arguments. The first methodology is based on natural language processing or computational linguistics. Given textual discourse, the goal involves identifying individual arguments, their internal structures and their interactions \cite{palau:09,lawrence:16}. Argumentation mining \cite{lippi:15,lippi:16,bar-haim2,zhao,bar-haim,toledo-ronen,boltuzic,cabrio,mayer:18,saint-dizier:18}, recognizing textual entailment \cite{zhao,silvia,levesque} and natural language inference \cite{bowman,maccartney} all belong to the first methodology. The first methodology is regarded successful if a conflict relation obtained with the approach conforms to a human judgement. The second methodology is based on the acceptability semantics \cite{dung:95} that emerged from the study of non-monotonic reasoning \cite{prakken:01,bench-capon:07}. Given acceptability of arguments, the goal involves identifying an attack relation explaining the acceptability well in terms of the acceptability semantics. It is interesting to investigate whether a conflict relation obtained with the second methodology is successful in the sense of the goal of the first approach. However, it is not a fundamental requirement of the second methodology. The argumentation framework (AF for short) synthesis problem \cite{niskanen} based on realizability \cite{dunne:15}, the abstract structure learning \cite{riveret}, and enforcement \cite{niskanen:18} and the generative models of the abstract argumentation \cite{kido:17-1,kido:18-1} all belong to the second methodology.
\par
The second methodology is much less studied compared with the first one in spite of its importance. Indeed, if it is faithful to the acceptability semantics then its solution is normative in the sense that a rational agent ought to accept the solution. This is because the acceptability semantics itself is a normative theory of human cognition regarding acceptability of arguments.\footnote{If it is a descriptive theory telling us what one actually or psychologically believes then it should be studied in psychology, rather than logic, requiring an empirical approach for its correctness. It is meanwhile a descriptive theory of many non-monotonic logics, e.g., Reiter's default logic, Pollock's inductive defeasible logics and logic programming.} It tells us which arguments a rational agent ought to believe when the agent accepts a given attack relation. A faithful methodology thus discusses which attack relation the agent ought to believe when the agent accepts an observation on acceptability of arguments. However, it is more difficult than it seems, to predict an attack relation in a way that is faithful to the acceptability semantics. The acceptability of arguments we observe in practice is one's sentiment regarding agreement or disagreement on the arguments. It often involves uncertainty due to a variety of reasons such as lack of data, the presence of noise in observation, and the existence of multiple solutions. A probability theory thus is useful to deal with the uncertainty in a certain way.
\par
However, what is still not clear is how to give a probabilistic account of an uncertain solution in a way that is faithful to the acceptability semantics. In this paper, we use probability theory to give a formal representation of problem-independent uncertain argumentation-theoretic inference. It is very different from another interesting research direction using probability theory to give a formal representation of problem-dependent uncertain domain knowledge, e.g., \cite{nielsen:07,saha:04,timmer:15,vreeswijk:05,grabmair:10,bex:16}. From a semantic point of view, we for the first time distinguish a concept of the inverse problem from a direct problem of the theory of abstract argumentation (or just the abstract argumentation) \cite{dung:95}, as follows.
\begin{description}
\item[Direct problem] Given an attack relation between arguments, a direct problem aims to find sentiments regarding acceptability of the arguments defined in accordance with the acceptability semantics.
\item[Inverse problem] Given noisy sentiments regarding acceptability of arguments, an inverse problem aims to find an attack relation between the arguments explaining the sentiments well in terms of the acceptability semantics.
\end{description}
From an inferential point of view, we then give a probabilistic model capturing the acceptability semantics in a probabilistic way. It enhances the worth of the acceptability semantics with the simple view that an attack relation does not deterministically exist, but they are probabilistically distributed. We argue that probabilistic inference on the model deals with a solution concept that is more comprehensive than solutions to both of the direct and inverse problems. Indeed, we show that a deterministic (i.e., non-probabilistic) solution to an inverse problem is a special case of our Bayesian solution. In contrast to the deterministic solution, the Bayesian solution allows us to take into account a subjective beliefs on the existence of attack relations and to handle uncertainty to what extent each attack relation is likely to be the case. We also show that a deterministic solution to a direct problem is a special case of our Bayesian solution. In contrast to the deterministic solution, the Bayesian solution allows us to handle the uncertainty to what extent acceptability of arguments is likely to be true and to deal with the situation where acceptability of arguments is caused by multiple attack relations distributed probabilistically.
\par
%
%
%
The contributions of this paper are summarized as follows. First, to our best knowledge, this is the first paper introducing a concept of the inverse problem to the field of computational argumentation. The past two decades in the field of computational argumentation in AI (artificial intelligence) witnessed an intensive study on defining acceptable arguments given various kinds of argumentation frameworks, e.g., \cite{cayrol,amgoud:09,bench-capon:02,modgil,leite,verheij,caminada,dung:06,baroni:05,coste-marquis}. This paper places them on the direct problem and turns the spotlight onto the inverse direction. Second, this paper lays a foundation for making acceptability semantics data-driven. From the data-driven point of view, a weakness of the study of the abstract argumentation, and a symbolic AI in general, is a knowledge acquisition bottleneck that is a problem on how to acquire knowledge from data. In contrast to the direct problem, the input of the inverse problem is an unstructured data and thus available on the web, e.g., votes in various social networking services. It makes us easier to find a killer AI application of acceptability semantics.
\par
This paper is organized as follows. In Section 2, we introduce the inverse problem of the abstract argumentation as a counterpart of the direct problem. In Section 3, we give a probabilistic model to provide a Bayesian solution to both types of the problems. Sections 4 and 5 discuss correctness of the probabilistic model in both theoretical and empirical manners. Section 6 concludes with discussion.

\section{Abstract Argumentation Problems}
%
%
\subsection{Direct Problems}
An abstract argumentation framework (AF) \cite{dung:95} is a pair $\langle arg, att\rangle$, where $arg$ denotes a set of arguments and $att$ denotes a binary relation on $arg$. $att$ represents an attack relation between arguments, i.e., $(a, b)\in att$ means ``$a$ attacks $b$.'' Suppose $a\in arg$ and $S\subseteq arg$. $S$ attacks $a$ if, and only if (iff), some member of $S$ attacks $a$. $S$ is conflict-free iff $S$ attacks none of its members. $a$ is acceptable with respect to $S$ iff $S$ attacks all arguments that attack $a$. A characteristic function $F: Pow(arg)\rightarrow Pow(arg)$ is defined as $F(S)=\{a | a$ is acceptable with respect to $S\}$ where $Pow(arg)$ is the power set of $arg$. $S$ is admissible iff $S$ is conflict-free and every member of $S$ is acceptable with respect to $S$. The acceptability semantics \cite{dung:95} defines four types of extensions of $AF$ that intuitively represent sets of acceptable arguments.
%
\begin{itemize}
\item A \emph{preferred extension} is a maximal (with respect to set inclusion) admissible set.
\item A conflict-free set $S$ of arguments is a \emph{stable extension} iff $S$ attacks each argument which does not belong to $S$.
\item The \emph{grounded extension} is the least fixed point of $F$.
\item An admissible set $S$ of arguments is a \emph{complete extension} iff each argument, which is acceptable with respect to $S$, belongs to $S$.
\end{itemize}
\par
Let $\varepsilon$, $arg$ and $att$ represent an acceptability semantics, a set of arguments, an attack relation on $arg$, respectively, and $acc$ represent the set of extensions of the argumentation framework, $\langle arg,att\rangle$, with respect to $\varepsilon$. When we see $\varepsilon$ as a function, the following equation holds.
\begin{eqnarray}\label{eq:af1}
\varepsilon(arg, att)=acc
\end{eqnarray}
Equation (\ref{eq:af1}) thus shows the relationship among a knowledge representation, i.e., $\langle arg, att\rangle$, a consequence, i.e., $acc$, and a semantics, i.e., $\varepsilon$. We assume that $arg$ and $\varepsilon$ are arbitrary but fixed. We define a direct (or forward) problem of the abstract argumentation, as follows.
%
%
%
\begin{definition}[Direct problem]
A direct problem of the abstract argumentation is defined as follows: Given an attack relation $att$, find an  acceptability $acc$ satisfying $acc=\varepsilon(arg,att)$.
\end{definition}
%

\subsection{Inverse Problems}
Next, we consider an inverse of the direct problem. In contrast to the direct problem assuming the existence of an attack relation, an inverse problem aims to determine an attack relation. It is natural to assume that some of an attack relation, denoted by $att^{\mathrm{k}}$, is known. The following equation thus holds from Equation (\ref{eq:af1}).
%
%
\begin{eqnarray}\label{eq:af2}
acc=\varepsilon(arg, att^{\mathrm{k}},att)
\end{eqnarray}
We again suppose that $arg$ and $\varepsilon$, and $att^{\mathrm{k}}$ as well, are arbitrary but fixed. We define an inverse problem of the abstract argumentation, as follows.\footnote{It is possible to think of another inverse problem where a semantics is also unknown, for instance. This paper, however, does not discuss it because it is not very practical but rather complicated.}
\begin{definition}[Inverse problem]\label{def:ip}
An inverse problem of the abstract argumentation is defined as follows: Given an acceptability $acc$, find an attack relation $att$ satisfying $acc=\varepsilon(arg, att^{\mathrm{k}},att)$.
\end{definition}
\par
A problem, either inverse or direct, is said to be well-posed if a solution exists, the solution is unique if it exists, and the solution depends continuously on the input, i.e., \emph{solution existence}, \emph{solution uniqueness} and \emph{solution stability}, respectively \cite{aster}. Neither an inverse nor direct problem of the abstract argumentation is well-posed, i.e., ill-posed. Indeed, the solution stability does not hold because they are a discrete problem rather than a continuous problem. The solution existence holds in a direct problem when semantics is not stable. However, it holds in neither direct nor inverse problems in general.
%
\begin{example}
Let $arg=\{a,b\}$ and $att^{\mathrm{k}}=\emptyset$. Given $acc=\{\emptyset,\{a,b\}\}$, there is no attack relation that is a solution to the inverse problem.
\end{example}
An acceptability is observed in an inverse problem. The solution existence does not generally hold in an inverse problem because it is empirically true that an observation often includes some amount of noise. Noise can be an effect irrelevant to semantics $\varepsilon$, or can be a false or inaccurate observation. A realistic inverse problem thus needs to find an attack relation $att$ satisfying the following equation:
\begin{eqnarray}\label{eq:af3}
acc=\varepsilon(arg, att^{\mathrm{k}},att)+\eta,
\end{eqnarray}
where $\eta$ and $+$ denote a noise and a \emph{notional} operator for addition, respectively. The presence of noise makes an inverse problem difficult because it requires a solution to distinguish the true acceptability from a noise inseparably observed.
\par
The solution uniqueness holds in a direct problem. It however does not hold in an inverse problem.
%
\begin{example}
Let $arg=\{a,b,c\}$ and $att^{\mathrm{k}}=\emptyset$. Given $acc=\{\{a,c\}\}$, the attack relations represented with the directed graphs below are all solutions to the inverse problem.
\begin{center}
 \includegraphics[scale=0.52]{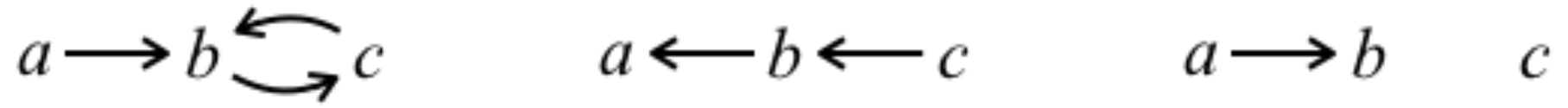}
\end{center}
\end{example}
There is a restriction on an attack relation and semantics that guarantees the solution uniqueness.
\begin{proposition}[Solution uniqueness]\label{prop:1}
An inverse problem satisfies the solution uniqueness if both known and unknown attack relations are symmetric and irreflexive, and semantics is either complete, preferred or stable.
\end{proposition}
The restriction of Proposition \ref{prop:1} is not a very practical nature. In practice, an observation consists of sentiments regarding acceptability of arguments, collected from a lot of individuals. Such sentiments would be explained well in terms of complete semantics because they are likely to include both credulous and skeptical sentiments. Moreover, self-attacking arguments rarely occur in practice and it is more useful to find the existence of attack between arguments rather than the direction of the attack.
%
\par
When a problem is ill-posed, it does not mean that the problem is inappropriate at all. It implies that its solution is inherently uncertain and it is effective to use a probability theory to formulate the uncertainty in a certain way. Moreover, there is another reason why a probabilistic approach is preferable. Almost all of the real arguments are an \emph{enthymeme}, i.e., an argument whose premise or conclusion is unexpressed. The existence of an attack relation between enthymemes depends on the contexts of argumentation and knowledge of the arguers and listeners. Let us take a look at a simple example of the uncertainty of the existence of an attack relation.
%
\begin{example}
Suppose one hundred individuals asked to express their opinions whether there is an attack between the following arguments $a$ and $b$.
\begin{itemize}
\item $a$: Tweety can fly because it is a bird.
\item $b$: Tweety is a penguin.
\end{itemize}
Many individuals would find an attack between them because they know that penguins cannot fly. However, not all individuals would find the attack because some individuals do not know that penguins cannot fly or some other individuals might know that Tweety is a genetically-altered flying penguin. Now, let us assume that ninety individuals think that there is an attack between them and the remaining ten individuals think that there is no attack. The best we can conclude is that the probability that there is an attack between them is 0.9.
\end{example}
The uncertainty of the existence of an attack relation leads to the idea that acceptability of arguments depends on multiple different attack relations probabilistically distributed. An inverse problem in this situation wants to find $N$ attack relations $att_{n}$ $(1\leq n\leq N)$ satisfying the following equation.
\begin{eqnarray}\label{eq:af4}
acc=\varepsilon(arg, att^{\mathrm{k}},att_{1})+\cdots+\varepsilon(arg, att^{\mathrm{k}},att_{N})+\eta
\end{eqnarray}
Meanwhile, the direct problem in this situation wants to find $acc$ given all $att_{n}$ $(1\leq n\leq N)$. Here, it is natural to think that different $att_{n}$ have different influences on $acc$. A probabilistic approach makes it easier to deal with the situation because it gives a formal account of the idea that an attack relation is probabilistically distributed. It is theoretically a general case of the view that an attack relation deterministically exists. We will discuss comprehensive solutions to both the inverse and direct problems later in Equation (\ref{eq:attackestimate}) and Equation (\ref{eq:8}), respectively.

%
\section{Abstract Argumentation Model}
This section gives a Bayesian account of direct and inverse problems of the abstract argumentation. It allows us to probabilistically deal with all of the situations, from (\ref{eq:af1}) to (\ref{eq:af4}), discussed in the previous section.
%
\subsection{Probabilistic Model for Abstract Argumentation Problems}
Let $arg$ be a set of arguments. We assume two kinds of random variables. For all $m\in arg\times arg$, $Att_{m}$ is a random variable representing the existence of an attack relation from its left to right elements of $m$, and $att_{m}$ represents a value of $Att_{m}$, either 0 or 1.\footnote{For the sake of simplicity, $m$ also represents a set of two arguments. $Att_{m}$ in this case represents the existence of a symmetric attack relation between the arguments in $m$.} For all $d\subseteq arg$, $Acc_{d}$ is a random variable representing acceptability of $d$, and $acc_{d}$ represent a value of $Acc_{d}$. $\bm{Att}$ and $\bm{att}$ are assumed to denote sequences of $Att_{m}$ and $att_{m}$, respectively. Similarly, $\bm{Acc}$ and $\bm{acc}$ denote sequences of $Acc_{d}$ and $acc_{d}$, respectively. Thus $\bm{att}$ and $\bm{acc}$ represent an attack relation on $arg$ and acceptability of $arg$, respectively. In this paper, we study the relationship between $\bm{Att}$ and $\bm{Acc}$. We thus assume that $arg$ and acceptability semantics are arbitrary but fixed.
%
\par
It is natural to assume that there are two cases, either $a$ attacks $b$ or $a$ does not attack $b$, i.e., $Att_{(a,b)}=1$ or $Att_{(a,b)}=0$, given a pair $(a,b)$ of arguments $a$ and $b$. It is moreover natural to assume that if the probability that $a$ attacks $b$ is $\lambda_{(a,b)}$ then the probability that $a$ does not attack $b$ is $1-\lambda_{(a,b)}$. We thus define the probability distribution of an attack-relation variable as follows.
\begin{definition}[Attack distribution]\label{def:attd}
Let $Att_{m}$ be a random variable of an attack relation and $\lambda_{m}$ be a constant such that $0\leq\lambda_{m}\leq 1$. The probability distribution over $Att_{m}$, denoted by $p(Att_{m})$, is given by
\begin{eqnarray*}
p(Att_{m})=\lambda_{m}^{Att_{m}}(1-\lambda_{m})^{1-Att_{m}}.
\end{eqnarray*}
\end{definition}
It is obvious that $p(Att_{m})=\lambda_{m}$ if $Att_{m}=1$ and $p(Att_{m})=1-\lambda_{m}$ if $Att_{m}=0$. An actual value of $\lambda_{m}$ is assumed to be given in advance by a subjective belief. Technically speaking, the distribution is called a \emph{Bernoulli distribution} \cite{bernoulli} often used to represent a discrete probability distribution that takes two values, 0 or 1.
\par
It is natural to assume that there are two cases, either a set $d$ of arguments is acceptable or not, denoted by $Acc_{d}=1$ or $Acc_{d}=0$, respectively. In terms of the acceptability semantics \cite{dung:95}, $Acc_{d}$ depends on $\bm{att}$, i.e., an attack relation on $arg$. Let us assume a constant $\theta_{d|\bm{att}}$ ($0\leq \theta_{d|\bm{att}}\leq 1$) representing the probability that the acceptability semantics makes $d$ acceptable given $\bm{att}$. The probability distribution over $Acc_{d}$ given $\bm{att}$ can be defined using a Bernoulli distribution with parameter $\theta_{d|\bm{att}}$.
\begin{definition}[Acceptability distribution]
Let $Acc_{d}$ be an acceptability variable, $\bm{att}$ be a sequence of values of attack-relation variables and $\theta_{d|\bm{att}}$ be a constant satisfying $0\leq \theta_{d|\bm{att}}\leq 1$. The probability distribution over $Acc_{d}$ given $\bm{att}$, denoted by $p(Acc_{d}|\bm{att})$, is given by
\begin{eqnarray*}
p(Acc_{d}|\bm{att})=\theta_{d|\bm{att}}^{Acc_{d}}(1-\theta_{d|\bm{att}})^{1-Acc_{d}}.
\end{eqnarray*}
\end{definition}
\par
$\theta_{d|\bm{att}}$ should take a high value when $\bm{att}$ explains the acceptability of $d$ well in terms of the acceptability semantics. It thus holds when there is an extension $e$ of the abstract argumentation framework of $\bm{att}$ such that there is a large agreement between $e$ and $d$. The agreement can be defined in terms of the following two aspects: a true positive and a true negative, denoted by $tp$ and $tn$, respectively.
\begin{eqnarray*}
tp(e,d)&=&\{a\in arg|a\in e, a\in d\}\\
tn(e,d)&=&\{a\in arg|a\notin e, a\notin d\}
\end{eqnarray*}
It is thus reasonable to assume that $\theta_{d|\bm{att}}$ depends on the cardinality of the union of the true positive and the true negative, i.e., $|tp(e,d)|+|tn(e,d)|$. We give three definitions of the acceptability parameter. The first one is a \emph{deterministic acceptability parameter}.
%
\begin{definition}[Deterministic acceptability parameter]
The deterministic acceptability parameter $\theta_{d|\bm{att}}$ is given by
\begin{eqnarray*}
\theta_{d|\bm{att}}=
\begin{cases}
1 & d\in\varepsilon(arg,\bm{att})\\
0 & otherwise.
\end{cases}
\end{eqnarray*}
\end{definition}
Therefore, $\theta_{d|\bm{att}}=1$ holds if and only if there is an extension $e$ of the argumentation framework of $\bm{att}$ such that $|tp(e,d)|+|tn(e,d)|=|arg|$ holds.
\par
The second parameter is a \emph{linear acceptability parameter} defined using a linear function $g(x)=wx+a$. Since $g$ is a monotonically increasing function, it is reasonable to define the acceptability parameters with the output of $g$ given $|tp(e,d)|+|tn(e,d)|$, i.e., $g(|tp(e,d)|+|tn(e,d)|)$. We thus normalize $g$ so that its domain and range become $[0,|arg|]$ and $[0,1]$, respectively. We then obtain
\begin{eqnarray*}
f(x)=\frac{g(x)-\min\{g(x)\}}{\max\{g(x)\}-\min\{g(x)\}}=\frac{wx+a-a}{w|arg|+a-a}=\frac{x}{|arg|}.
\end{eqnarray*}
%
We define $\theta_{d|\bm{att}}$ as $f(|tp(e,d)|+|tn(e,d)|)$ where $|tp(e,d)|+|tn(e,d)|$ is maximum with respect to an extension $e$ of the argumentation framework of $\bm{att}$.
\begin{definition}[Linear acceptability parameter]\label{def:LAP}
The linear acceptability parameter $\theta_{d|\bm{att}}$ is given by
\begin{eqnarray*}
\theta_{d|\bm{att}}=\frac{1}{|arg|}\max\left\{|tp(d,e)|+|tn(d,e)|\mathrel{\Big|}e\in\varepsilon(arg,\bm{att})\right\}.
\end{eqnarray*}
\end{definition}
$\theta_{d|\bm{att}}$ thus increases $1/|arg|$ when another argument becomes a member of the true positive or true negative.
\par
The third parameter is an \emph{exponential acceptability parameter} defined using an exponential function $g(x)=a(w)^{x}$. Since $g$ is a monotonically increasing function given $w\geq 1$ and $a\geq 0$, it is similarly reasonable to define the acceptability parameters with $g(|tp(e,d)|+|tn(e,d)|)$. We thus normalize $g$ so that its domain and range become $[0,|arg|]$ and $[0,1]$, respectively. We then obtain
\begin{eqnarray}
f(x)=\frac{g(x)-\min\{g(x)\}}{\max\{g(x)\}-\min\{g(x)\}}=\frac{a(w)^{x}-a(w)^{0}}{a(w)^{|arg|}-a(w)^{0}}=\frac{w^{x}-1}{w^{|arg|}-1}.\label{eq:nef}
\end{eqnarray}
We similarly define $\theta_{d|\bm{att}}$ as $f(|tp(e,d)|+|tn(e,d)|)$ where $|tp(e,d)|+|tn(e,d)|$ is maximum with respect to an extension $e$ of the argumentation framework of $\bm{att}$.
%
\begin{definition}[Exponential acceptability parameter]\label{def:eap}
Let $w> 1$. The exponential acceptability parameter $\theta_{d|\bm{att}}$ is given by
\begin{eqnarray*}
\theta_{d|\bm{att}}=\frac{1}{w^{|arg|}-1}\max\left\{w^{|tp(d,e)|+|tn(d,e)|}-1\mathrel{\Big|}e\in\varepsilon(arg,\bm{att})\right\}.
\end{eqnarray*}
\end{definition}
%
Given a large value $w$, $\theta_{d|\bm{att}}$ approximates $w^{|tp(d,e)|+|tn(d,e)|}/w^{|arg|}$. In this case, $\theta_{d|\bm{att}}$ increases $w$ times when another argument becomes a member of the true positive or true negative.
\par
The exponential acceptability parameter has good properties. First, the deterministic acceptability parameter is a special case of the exponential acceptability parameter.
\begin{proposition}\label{prop:2}
Let $\theta_{d|\bm{att}}$ be an exponential acceptability parameter and $\phi_{d|\bm{att}}$ be a deterministic acceptability parameter. $\lim_{w\to\infty}\theta_{d|\bm{att}}=\phi_{d|\bm{att}}$ holds.
\end{proposition}
%
\par
Second, the linear acceptability parameter is also a special case of the exponential acceptability parameter.
\begin{proposition}\label{prop:3}
Let $\theta_{d|\bm{att}}$ be an exponential acceptability parameter and $\phi_{d|\bm{att}}$ be a linear acceptability parameter. $\lim_{w\to1}\theta_{d|\bm{att}}=\phi_{d|\bm{att}}$ holds.
\end{proposition}
\par
\begin{figure}[t]
\begin{center}
 \includegraphics[scale=0.6]{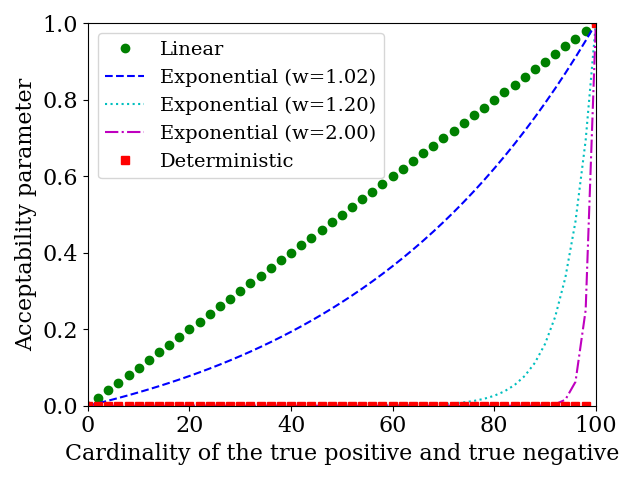}
  \caption{Acceptability parameters versus the cardinality of the true positive and true negative, given one hundred arguments.}
  \label{fig:parameters}
\end{center}
\end{figure}

Figure \ref{fig:parameters} shows each type of the acceptability parameter. It is visually shown that both of the deterministic and linear acceptability parameters are extreme cases of the exponential acceptability parameter.
%
\begin{example}
Table \ref{ex.theta} shows all possible linear and exponential acceptability parameters given $arg=\{a,b,c\}$ and $\bm{att}=(att_{\{a,b\}}, att_{\{a,c\}}, att_{\{b,c\}})$. It is observed that the exponential acceptability parameters give a relatively sharp distribution compared with the linear acceptability parameters.
{\small
\begin{table*}[t]
\caption{Linear acceptability parameters (upper) and exponential acceptability parameters (lower) defined with complete semantics. $\theta_{X}$ is an abbreviation for $\theta_{X|\bm{att}}$ where $\bm{att}=(att_{\{a,b\}},att_{\{a,c\}},att_{\{b,c\}})$.}
\label{ex.theta}
\begin{center}
\begin{tabular}{c|cccccccc}
$\bm{att}$ & $\theta_{\emptyset}$ & $\theta_{\{a\}}$ & $\theta_{\{b\}}$ & $\theta_{\{c\}}$ & $\theta_{\{a,b\}}$ & $\theta_{\{a,c\}}$ & $\theta_{\{b,c\}}$ & $\theta_{\{a,b,c\}}$\\\hline\hline
$(0,0,0)$ & $0$ & $1/3$ & $1/3$ & $1/3$ & $2/3$ & $2/3$ & $2/3$ & $1$\\
$(1,0,0)$ & $2/3$ & $2/3$ & $2/3$ & $1$ & $1/3$ & $1$ & $1$ & $2/3$\\ 
$(0,1,0)$ & $2/3$ & $2/3$ & $1$ & $2/3$ & $1$ & $1/3$ & $1$ & $2/3$\\
$(1,1,0)$ & $1$ & $1$ & $2/3$ & $2/3$ & $2/3$ & $2/3$ & $1$ & $2/3$\\
$(0,0,1)$ & $2/3$ & $1$ & $2/3$ & $2/3$ & $1$ & $1$ & $1/3$ & $2/3$\\
$(1,0,1)$ & $1$ & $2/3$ & $1$ & $2/3$ & $2/3$ & $1$ & $2/3$ & $2/3$\\
$(0,1,1)$ & $1$ & $2/3$ & $2/3$ & $1$ & $1$ & $2/3$ & $2/3$ & $2/3$\\
$(1,1,1)$ & $1$ & $1$ & $1$ & $1$ & $2/3$ & $2/3$ & $2/3$ & $1/3$\\\hline
$(0,0,0)$ & $0$ & $\frac{w-1}{w^{3}-1}$ & $\frac{w-1}{w^{3}-1}$ & $\frac{w-1}{w^{3}-1}$ & $\frac{w^{2}-1}{w^{3}-1}$ & $\frac{w^{2}-1}{w^{3}-1}$ & $\frac{w^{2}-1}{w^{3}-1}$ & $1$\\
$(1,0,0)$ & $\frac{w^{2}-1}{w^{3}-1}$ & $\frac{w^{2}-1}{w^{3}-1}$ & $\frac{w^{2}-1}{w^{3}-1}$ & $1$ & $\frac{w-1}{w^{3}-1}$ & $1$ & $1$ & $\frac{w^{2}-1}{w^{3}-1}$\\ 
$(0,1,0)$ & $\frac{w^{2}-1}{w^{3}-1}$ & $\frac{w^{2}-1}{w^{3}-1}$ & $1$ & $\frac{w^{2}-1}{w^{3}-1}$ & $1$ & $\frac{w-1}{w^{3}-1}$ & $1$ & $\frac{w^{2}-1}{w^{3}-1}$\\
$(1,1,0)$ & $1$ & $1$ & $\frac{w^{2}-1}{w^{3}-1}$ & $\frac{w^{2}-1}{w^{3}-1}$ & $\frac{w^{2}-1}{w^{3}-1}$ & $\frac{w^{2}-1}{w^{3}-1}$ & $1$ & $\frac{w^{2}-1}{w^{3}-1}$\\
$(0,0,1)$ & $\frac{w^{2}-1}{w^{3}-1}$ & $1$ & $\frac{w^{2}-1}{w^{3}-1}$ & $\frac{w^{2}-1}{w^{3}-1}$ & $1$ & $1$ & $\frac{w-1}{w^{3}-1}$ & $\frac{w^{2}-1}{w^{3}-1}$\\
$(1,0,1)$ & $1$ & $\frac{w^{2}-1}{w^{3}-1}$ & $1$ & $\frac{w^{2}-1}{w^{3}-1}$ & $\frac{w^{2}-1}{w^{3}-1}$ & $1$ & $\frac{w^{2}-1}{w^{3}-1}$ & $\frac{w^{2}-1}{w^{3}-1}$\\
$(0,1,1)$ & $1$ & $\frac{w^{2}-1}{w^{3}-1}$ & $\frac{w^{2}-1}{w^{3}-1}$ & $1$ & $1$ & $\frac{w^{2}-1}{w^{3}-1}$ & $\frac{w^{2}-1}{w^{3}-1}$ & $\frac{w^{2}-1}{w^{3}-1}$\\
$(1,1,1)$ & $1$ & $1$ & $1$ & $1$ & $\frac{w^{2}-1}{w^{3}-1}$ & $\frac{w^{2}-1}{w^{3}-1}$ & $\frac{w^{2}-1}{w^{3}-1}$ & $\frac{w-1}{w^{3}-1}$
\end{tabular}
\end{center}
\end{table*}
}
\end{example}
For the sake of generality, we assume the exponential acceptability parameter unless otherwise stated. All of the acceptability parameters are \emph{faithful} to acceptability semantics in the sense that no heuristic is introduced in their definitions. The exponential acceptability parameter is indeed just a normalization of an exponential function and the linear acceptability parameter is also just a normalization of a linear function.
\par
\begin{figure}[t]
\begin{center}
 \includegraphics[scale=0.3]{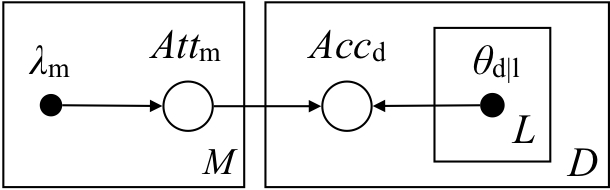}
  \caption{Graphical representation of the abstract argumentation model.}
  \label{argumentationmodel}
\end{center}
\end{figure}

Figure \ref{argumentationmodel} shows a graphical representation of dependencies of the random variables and deterministic parameters we introduced in this section. The boxes, so-called plates, represent that there are $M$ nodes of $Att_{m}$ and $\lambda_{m}$ for each pair $m$ of arguments, $D$ nodes of $Acc_{d}$ for each set $d$ of arguments, and $L\times D$ nodes of $\theta_{d|l}$ for each pair of $d$ and a possible attack relation $\bm{att}$ labelled $l$. We call the components of the Bayesian network an \emph{abstract argumentation model} and represent it with ${\cal M}$.
\par
Theoretically speaking, it is a \emph{mixture model} in the sense that the parent distribution for $Att_{m}$ influences the child distribution of $Acc_{d}$. It is however different from a custom, e.g., mixtures of Bernoulli distributions or Gaussian distributions \cite{bishop}. They assume that each value of a child variable can be generated from a different value of a parent variable. By contract, the abstract argumentation model assumes that every acceptability is generated from the same attack relations. This comes from our basic assumption that an attack relation is statistically an objective opinion although acceptability of arguments is a subjective opinion determined by an extension subjectively chosen by individuals.
%
\subsection{Probabilistic Inference of Attack Relations}
Abstract argumentation model ${\cal M}$ is a \emph{generative model} in the sense that it gives a formal account of the argumentation-theoretic causality, how acceptability of a set of arguments is interpreted given an argumentation framework in accordance with acceptability semantics. In an inverse problem, we use the model to trace the causality back to the argumentation framework from acceptability. Technically speaking, this is achieved by calculating the posterior distribution over attack relations given values of acceptability variables, i.e., $p(\bm{Att}|\bm{acc})$. Using Bayes' theorem, we obtain
%
\begin{eqnarray}
p(\bm{Att}|\bm{acc})&=&\frac{p(\bm{acc}|\bm{Att})p(\bm{Att})}{p(\bm{acc})}\nonumber\\
&\propto&p(\bm{acc}|\bm{Att})p(\bm{Att})\nonumber\\
&=&\prod_{d}^{D}p(acc_{d}|\bm{Att})\prod_{m}^{M}p(Att_{m})\nonumber\\
&=&\prod_{d}^{D}\theta_{d|\bm{Att}}^{acc_{d}}(1-\theta_{d|\bm{Att}})^{1-acc_{d}}\prod_{m}^{M}\lambda_{m}^{Att_{m}}(1-\lambda_{m})^{1-Att_{m}}\label{eq:attackestimate}
\end{eqnarray}
where $x\propto y$ means ``$x$ is proportional to $y$'' and thus there is a constant $K$ such that $x=Ky$ holds. In ${\cal M}$, each acceptability is independently distributed from the same distributions over attack relations. This property is said to be \emph{independent and identically distributed, i.e., i.i.d.}. It results in the desirable property that enables to successively update the posterior distributions over attack relations whenever an acceptability is observed. In fact, it is obvious from Expression (\ref{eq:attackestimate}) that we have the following equation when observing $\bm{acc}$, values of acceptability variables.
%
%
\begin{eqnarray*}
p(\bm{Att}|\bm{acc})&\propto&p(\bm{Att})\prod_{d}^{D}p(acc_{d}|\bm{Att})
\end{eqnarray*}
When another acceptability $acc_{D+1}$ is observed, the above equation leads to the result
%
\begin{eqnarray*}
p(\bm{Att}|\bm{acc},acc_{D+1})&\propto&p(\bm{Att})\prod_{d}^{D+1}p(acc_{d}|\bm{Att})\\
&=&p(\bm{Att})p(acc_{D+1}|\bm{Att})\prod_{d}^{D}p(acc_{d}|\bm{Att})\\
&\propto&p(\bm{Att}|\bm{acc})p(acc_{D+1}|\bm{Att}).
\end{eqnarray*}
Therefore, the most recent posterior distribution is proportional to the product of the previous posterior distribution and the likelihood of the new observation.
%
\begin{example}
We here see how the probability distribution over attack relations is updated one by one. Let us assume set $arg=\{a,b,c\}$ of arguments and three random variables $Att_{\{a,b\}}$, $Att_{\{a,c\}}$, $Att_{\{b,c\}}$ of an attack relation. Now, suppose the observation that set $\{a\}$ is acceptable, i.e., $Acc_{\{a\}}=1$. The posterior distribution over attack-relation variables given the observation is represented by
%
%
%
\begin{eqnarray*}
&&p(Att_{\{a,b\}},Att_{\{a,c\}},Att_{\{b,c\}}|Acc_{\{a\}}=1)\\
&\propto&p(Acc_{\{a\}}=1|Att_{\{a,b\}},Att_{\{a,c\}},Att_{\{b,c\}})p(Att_{\{a,b\}})p(Att_{\{a,c\}})p(Att_{\{b,c\}})\\
&=&\theta_{\{a\}|Att_{\{a,b\}},Att_{\{a,c\}},Att_{\{b,c\}}}\prod_{m\in\{\{a,b\},\{a,c\},\{b,c\}\}}\lambda_{m}^{Att_{m}}(1-\lambda_{m})^{1-Att_{m}}.
\end{eqnarray*}
Let $\bm{acc}$ be $(Acc_{\{a\}}=1)$ and $\theta_{d|\bm{att}}$ be the exponential acceptability prior shown in Table \ref{ex.theta} defined with complete semantics. We here suppose $w=2$. The posterior distribution is given as follows.
%
%
\begin{eqnarray*}
p(0,0,0|\bm{acc})&\propto&(1/7)(1-\lambda_{\{a,b\}})(1-\lambda_{\{a,c\}})(1-\lambda_{\{b,c\}})\\
p(1,0,0|\bm{acc})&\propto&(3/7)\lambda_{\{a,b\}}(1-\lambda_{\{a,c\}})(1-\lambda_{\{b,c\}})\\
p(0,1,0|\bm{acc})&\propto&(3/7)(1-\lambda_{\{a,b\}})\lambda_{\{a,c\}}(1-\lambda_{\{b,c\}})\\
p(1,1,0|\bm{acc})&\propto&\lambda_{\{a,b\}}\lambda_{\{a,c\}}(1-\lambda_{\{b,c\}})\\
p(0,0,1|\bm{acc})&\propto&(1-\lambda_{\{a,b\}})(1-\lambda_{\{a,c\}})\lambda_{\{b,c\}}\\
p(1,0,1|\bm{acc})&\propto&(3/7)\lambda_{\{a,b\}}(1-\lambda_{\{a,c\}})\lambda_{\{b,c\}}\\
p(0,1,1|\bm{acc})&\propto&(3/7)(1-\lambda_{\{a,b\}})\lambda_{\{a,c\}}\lambda_{\{b,c\}}\\
p(1,1,1|\bm{acc})&\propto&\lambda_{\{a,b\}}\lambda_{\{a,c\}}\lambda_{\{b,c\}}
%
\end{eqnarray*}
%
%
Next, we suppose another observation that set $\{b\}$ is acceptable, i.e., $Acc_{\{b\}}=1$. The posterior distribution is updated as follows in accordance with the update equation.
%
\begin{eqnarray*}
&&p(Att_{\{a,b\}},Att_{\{a,c\}},Att_{\{b,c\}}|Acc_{\{a\}}=1,Acc_{\{b\}}=1)\\
&\propto&p(Att_{\{a,b\}},Att_{\{a,c\}},Att_{\{b,c\}}|Acc_{\{a\}}=1)p(Acc_{\{b\}}=1|Att_{\{a,b\}},Att_{\{a,c\}},Att_{\{b,c\}})\\
&\propto&p(Att_{\{a,b\}},Att_{\{a,c\}},Att_{\{b,c\}}|Acc_{\{a\}}=1)\theta_{\{b\}|Att_{\{a,b\}},Att_{\{a,c\}},Att_{\{b,c\}}}
\end{eqnarray*}
Now, let $\bm{acc}$ be $(Acc_{\{a\}}=1,Acc_{\{b\}}=1)$. The posterior distribution is updated as follows.
%
\begin{eqnarray*}
p(0,0,0|\bm{acc})&\propto&(1/7)^{2}(1-\lambda_{\{a,b\}})(1-\lambda_{\{a,c\}})(1-\lambda_{\{b,c\}})\\
p(1,0,0|\bm{acc})&\propto&(3/7)^{2}\lambda_{\{a,b\}}(1-\lambda_{\{a,c\}})(1-\lambda_{\{b,c\}})\\
p(0,1,0|\bm{acc})&\propto&(3/7)(1-\lambda_{\{a,b\}})\lambda_{\{a,c\}}(1-\lambda_{\{b,c\}})\\
p(1,1,0|\bm{acc})&\propto&(3/7)\lambda_{\{a,b\}}\lambda_{\{a,c\}}(1-\lambda_{\{b,c\}})\\
p(0,0,1|\bm{acc})&\propto&(3/7)(1-\lambda_{\{a,b\}})(1-\lambda_{\{a,c\}})\lambda_{\{b,c\}}\\
p(1,0,1|\bm{acc})&\propto&(3/7)\lambda_{\{a,b\}}(1-\lambda_{\{a,c\}})\lambda_{\{b,c\}}\\
p(0,1,1|\bm{acc})&\propto&(3/7)^{2}(1-\lambda_{\{a,b\}})\lambda_{\{a,c\}}\lambda_{\{b,c\}}\\
p(1,1,1|\bm{acc})&\propto&\lambda_{\{a,b\}}\lambda_{\{a,c\}}\lambda_{\{b,c\}}
\end{eqnarray*}
We further suppose an additional observation that set $\{c\}$ is acceptable, i.e., $Acc_{\{c\}}=1$. The posterior distribution is updated as follows.
%
\begin{eqnarray*}
&&p(Att_{\{a,b\}},Att_{\{a,c\}},Att_{\{b,c\}}|Acc_{\{a\}}=1,Acc_{\{b\}}=1,Acc_{\{c\}}=1)\\
&\propto&p(Att_{\{a,b\}},Att_{\{a,c\}},Att_{\{b,c\}}|Acc_{\{a\}}=1,Acc_{\{b\}}=1)\theta_{\{c\}|Att_{\{a,b\}},Att_{\{a,c\}},Att_{\{b,c\}}}
\end{eqnarray*}
Let $\bm{acc}$ be $(Acc_{\{a\}}=1,Acc_{\{b\}}=1, Acc_{\{c\}}=1)$. The posterior distribution is updated as follows.
%
\begin{eqnarray*}
p(0,0,0|\bm{acc})&\propto&(1/7)^{3}(1-\lambda_{\{a,b\}})(1-\lambda_{\{a,c\}})(1-\lambda_{\{b,c\}})\\
p(1,0,0|\bm{acc})&\propto&(3/7)^{2}\lambda_{\{a,b\}}(1-\lambda_{\{a,c\}})(1-\lambda_{\{b,c\}})\\
p(0,1,0|\bm{acc})&\propto&(3/7)^{2}(1-\lambda_{\{a,b\}})\lambda_{\{a,c\}}(1-\lambda_{\{b,c\}})\\
p(1,1,0|\bm{acc})&\propto&(3/7)^{2}\lambda_{\{a,b\}}\lambda_{\{a,c\}}(1-\lambda_{\{b,c\}})\\
p(0,0,1|\bm{acc})&\propto&(3/7)^{2}(1-\lambda_{\{a,b\}})(1-\lambda_{\{a,c\}})\lambda_{\{b,c\}}\\
p(1,0,1|\bm{acc})&\propto&(3/7)^{2}\lambda_{\{a,b\}}(1-\lambda_{\{a,c\}})\lambda_{\{b,c\}}\\
p(0,1,1|\bm{acc})&\propto&(3/7)^{2}(1-\lambda_{\{a,b\}})\lambda_{\{a,c\}}\lambda_{\{b,c\}}\\
p(1,1,1|\bm{acc})&\propto&\lambda_{\{a,b\}}\lambda_{\{a,c\}}\lambda_{\{b,c\}}
\end{eqnarray*}
\par
In general, we suppose that $Acc_{\{a\}}=1$, $Acc_{\{b\}}=1$ and $Acc_{\{c\}}=1$ are repeatedly observed $N$ times in total in this order. Given $N$ observations, denoted by $\bm{acc}$, the posterior distribution is given as follows.
%
\begin{eqnarray*}
p(0,0,0|\bm{acc})&\propto&(1/7)^{N}(1-\lambda_{\{a,b\}})(1-\lambda_{\{a,c\}})(1-\lambda_{\{b,c\}})\\
p(1,0,0|\bm{acc})&\propto&(3/7)^{N-\lfloor N/3\rfloor}\lambda_{\{a,b\}}(1-\lambda_{\{a,c\}})(1-\lambda_{\{b,c\}})\\
p(0,1,0|\bm{acc})&\propto&(3/7)^{N-\lfloor (N+1)/3\rfloor}(1-\lambda_{\{a,b\}})\lambda_{\{a,c\}}(1-\lambda_{\{b,c\}})\\
p(1,1,0|\bm{acc})&\propto&(3/7)^{N-\lfloor (N+2)/3\rfloor}\lambda_{\{a,b\}}\lambda_{\{a,c\}}(1-\lambda_{\{b,c\}})\\
p(0,0,1|\bm{acc})&\propto&(3/7)^{N-\lfloor (N+2)/3\rfloor}(1-\lambda_{\{a,b\}})(1-\lambda_{\{a,c\}})\lambda_{\{b,c\}}\\
p(1,0,1|\bm{acc})&\propto&(3/7)^{N-\lfloor (N+1)/3\rfloor}\lambda_{\{a,b\}}(1-\lambda_{\{a,c\}})\lambda_{\{b,c\}}\\
p(0,1,1|\bm{acc})&\propto&(3/7)^{N-\lfloor N/3\rfloor}(1-\lambda_{\{a,b\}})\lambda_{\{a,c\}}\lambda_{\{b,c\}}\\
p(1,1,1|\bm{acc})&\propto&\lambda_{\{a,b\}}\lambda_{\{a,c\}}\lambda_{\{b,c\}}
\end{eqnarray*}
Here, $\lfloor x\rfloor$ denotes the floor function that returns the maximum integer that is equal to or less than $x$. Figure \ref{fig:PD} shows the posterior distribution over attack relations versus the number of observations. When the number of observations increases, it is observed that the posterior probability of the attack relation, $(Att_{\{a,b\}}=1,Att_{\{a,c\}}=1,Att_{\{b,c\}}=1)$, converges to one. The result is reasonable in terms of complete semantics because the attack relation can successfully explain all of the observations.
%
%
\begin{figure}[t]
\begin{center}
\includegraphics[scale=0.6]{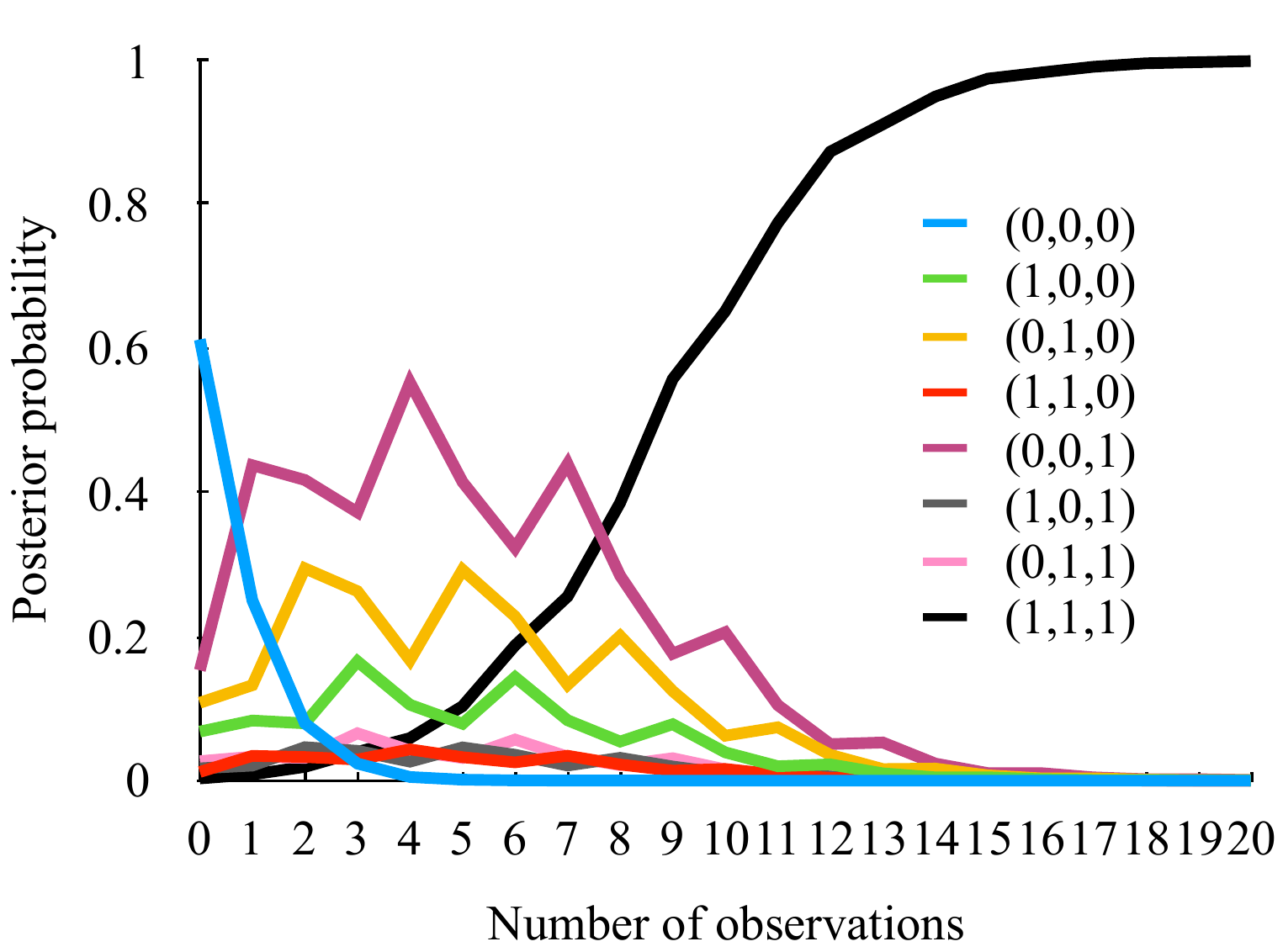}
  \caption{The horizontal axis shows the twenty iterated observations of $Acc_{\{a\}}=1$, $Acc_{\{b\}}=1$ and $Acc_{\{c\}}=1$ in this order. The vertical axis shows the posterior probability of each attack relation $(att_{\{a,b\}}, att_{\{a,c\}}, att_{\{b,c\}})$. We used the following parameters: $\lambda_{\{a,b\}}=0.1$, $\lambda_{\{a,c\}}=0.15$, $\lambda_{\{b,c\}}=0.2$ and the exponential acceptability parameters $\theta_{d|\bm{att}}$ shown in Table \ref{ex.theta} with $w=2$.}
  \label{fig:PD}
\end{center}
\end{figure}
\end{example}
\section{Theoretical Evaluation of Model Correctness}
\subsection{Comprehensive Solutions to Inverse Problems}
In this section, we investigate the relationship between model ${\cal M}$ and the inverse problem of the abstract argumentation. For the sake of simplicity, we do not distinguish $att\subseteq arg\times arg$ and the sequence $\bm{att}$ of values of attack-relation variables where $(Att_{(a,b)}=1)\in\bm{att}$ if $(a,b)\in att$ and $(Att_{(a,b)}=0)\in\bm{att}$ if $(a,b)\notin att$, for all arguments $a,b\in arg$. Similarly, we do not distinguish $acc\subseteq Pow(arg)$ and the sequence $\bm{acc}$ of values of acceptability variables where $(Acc_{d}=1)\in\bm{acc}$ if $d\in acc$ and $(Acc_{d}=0)\in\bm{acc}$ if $d\notin acc$, for all $d\in Pow(arg)$.
%
\par
An estimate with the maximum likelihood (so-called an \emph{ML estimation}) gives an attack relation that maximizes the likelihood of an observed acceptability. It has the form
\begin{eqnarray*}
\hat{\bm{att}}&=&\argmax_{\bm{att}}p(\bm{acc}|\bm{att}).
\end{eqnarray*}
The following theorem states that a solution to an inverse problem is an ML estimation in ${\cal M}$.
\begin{theorem}\label{thm:IPtoML}
Let $\bm{att}\subseteq arg\times arg$ and $\bm{acc}\subseteq Pow(arg)$. Given $\bm{acc}$, if $\bm{att}$ is a solution to the inverse problem then $\bm{att}$ is an ML estimation in ${\cal M}$.
\end{theorem}
\par
The converse of Theorem \ref{thm:IPtoML} does not hold in general.
\begin{theorem}\label{thm:MLtoIP}
Let $\bm{att}\subseteq arg\times arg$ and $\bm{acc}\subseteq Pow(arg)$. Given $\bm{acc}$, if $\bm{att}$ is an ML estimation in ${\cal M}$ then it is not necessarily true that $\bm{att}$ is a solution to the inverse problem.
\end{theorem}
\par
Theorems \ref{thm:IPtoML} and \ref{thm:MLtoIP} state that if it is a solution to an inverse problem then it is a solution to an ML estimation but not vice versa. This fact implies that a solution to an ML estimation is weaker than that of an inverse problem. However, the weakness does not mean worthless. In fact, the weakness of an ML estimation enables to deal with the presence of noise and the multiplicity of solutions we discussed with Equations (\ref{eq:af3}) and (\ref{eq:af4}), respectively.
\begin{example}
Given $arg=\{a,b\}$, let us suppose $\bm{acc}=\{\emptyset,\{a,b\}\}$ and $att^{\mathrm{k}}=\emptyset$. Given $\bm{acc}$, there is no solution to the inverse problem. However, $(Att_{(a,b)}=1,Att_{(b,a)}=1)$ is the ML estimation because we have
\begin{eqnarray*}
&&p(Acc_{\emptyset}=1,Acc_{\{a,b\}}=1|Att_{(a,b)}=0,Att_{(b,a)}=0)\propto\theta_{\emptyset|0,0}\theta_{\{a,b\}|0,0}=0\\
&&p(Acc_{\emptyset}=1,Acc_{\{a,b\}}=1|Att_{(a,b)}=1,Att_{(b,a)}=0)\propto\theta_{\emptyset|1,0}\theta_{\{a,b\}|1,0}=\frac{1}{9}\\
&&p(Acc_{\emptyset}=1,Acc_{\{a,b\}}=1|Att_{(a,b)}=0,Att_{(b,a)}=1)\propto\theta_{\emptyset|0,1}\theta_{\{a,b\}|0,1}=\frac{1}{9}\\
&&p(Acc_{\emptyset}=1,Acc_{\{a,b\}}=1|Att_{(a,b)}=1,Att_{(b,a)}=1)\propto\theta_{\emptyset|1,1}\theta_{\{a,b\}|1,1}=\frac{1}{3}.
\end{eqnarray*}
Here, we have assumed attack parameters $\lambda_{(a,b)}=\lambda_{(b,a)}=1/2$ and exponential acceptability parameter $\theta_{d|\bm{att}}$ with $w=2$ defined by complete semantics. This example shows that the use of an ML estimation is more useful than the deterministic solution because the deterministic solution can only tell us the fact that there is no solution to the inverse problem.
\end{example}
\par
We can benefit more from our probabilistic model ${\cal M}$ due to the theoretical fact that an ML estimation is an approximation of the Bayesian inference of $p(\bm{Att}|\bm{acc})$. Since it is a probability distribution, it is written as an $N$-tuple
\begin{eqnarray*}
p(\bm{Att}|\bm{acc})=\langle p(\bm{att}_{1}|\bm{acc}),p(\bm{att}_{2}|\bm{acc}),\cdots,p(\bm{att}_{N}|\bm{acc}) \rangle,
\end{eqnarray*}
where $N$ is the number of possible different attack relations. Now, we assume that the posterior distribution has a sharp peak at a unique attack relation $\hat{\bm{att}}$, i.e., $p(\bm{Att}|\bm{acc})=1$ if $\bm{Att}=\hat{\bm{att}}$ and otherwise $p(\bm{Att}|\bm{acc})=0$. Then, the Bayesian inference corresponds to solve the equation
\begin{eqnarray*}
\bm{\hat{att}}=\argmax_{\bm{att}}p(\bm{att}|\bm{acc})=\argmax_{\bm{att}}p(\bm{acc}|\bm{att})p(\bm{att}).
\end{eqnarray*}
%
In the last expression, we applied Bayes' theorem and eliminated the denominator irrelevant to the maximization. Interestingly, the equation represents an estimate with a maximum a posteriori (so-called an \emph{MAP estimation}). Next, we assume that the prior distribution $p(\bm{Att})$ over attack relations is uniform, i.e., constant $c$. We then have
%
%
\begin{eqnarray*}
\bm{\hat{att}}=\argmax_{\bm{att}}p(\bm{acc}|\bm{att})c=\argmax_{\bm{att}}p(\bm{acc}|\bm{att}).
\end{eqnarray*}
%
Constant $c$ is eliminated in the last expression because of its irrelevance to the maximization. Interestingly, this equation is an ML estimation. In sum, we saw that an ML estimation is obtained by restricting the prior distribution of an MAP estimation, and that an MAP estimation is obtained by restricting the posterior distribution of a Bayesian inference.
\par
The theoretical consequences discussed in this section are summarized as follows. First, an ML estimation is more flexible than a solution to an inverse problem. Second. an ML estimation is a special case of an MAP estimation. Third, an MAP estimation is a special case of a Bayesian inference. The practical implications of the consequences are summarized as follows.
\begin{itemize}
\item It is reasonable to use an ML estimation instead of a deterministic approach to solve an inverse problem because it gives a solution regardless of the existence of a noise in an observation.
\item It is reasonable to use an MAP estimation instead of an ML estimation because it allows us to reflect a subjective belief on attack relations.
\item It is reasonable to use a Bayesian inference instead of an MAP estimation because it tells us the uncertainty to what extent each attack relation is likely to be the case.
\end{itemize}
\par
\begin{figure*}
\begin{center}
\begin{tabular}{cc}
\begin{minipage}{0.49\hsize}
\begin{center}
 \includegraphics[scale=0.6]{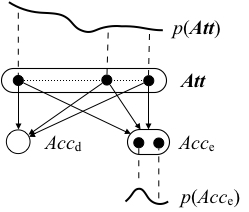}
\end{center}
\end{minipage}
&
\begin{minipage}{0.49\hsize}
\begin{center}
 \includegraphics[scale=0.6]{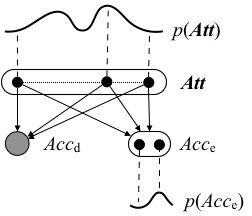}
\end{center}
\end{minipage}
\end{tabular}
\caption{From the point of view of the inverse problem, the two graphs illustrate that the observation of the acceptability of $d$, i.e., the grey node, updates the posterior attack distribution, i.e., the upper curve. From the point of view of the direct problem, the two graphs illustrate that the updated attack distribution updates the posterior acceptability distribution of $e$, i.e., the lower curve.
}
\label{fig:illustration}
\end{center}
\end{figure*}

Figure \ref{fig:illustration} illustrates an advantage of the use of a Bayesian inference in the inverse problem. It shows that attack relations are distributed probabilistically and its probability distribution is updated in accordance with an observed acceptability. A deterministic solution to the inverse problem is obviously its special case where there is only one attack relation whose posterior probability is one.

\subsection{Comprehensive Solutions to Direct Problems}
We next investigate the relationship between model ${\cal M}$ and the direct problem of the abstract argumentation. A prediction with a maximum likelihood (so-called an \emph{ML prediction}) gives an acceptability that is most likely to be generated from a given attack relation. It has the form
\begin{eqnarray*}
\hat{\bm{acc}}=\argmax_{\bm{acc}}p(\bm{acc}|\bm{att}).
\end{eqnarray*}
Thus, an acceptability, denoted by $\hat{\bm{acc}}$, given by an ML prediction maximizes its likelihood. The following relation exists between the direct problem and an ML prediction in ${\cal M}$.
\begin{theorem}\label{thm:MLandDP}
Let $\bm{att}\subseteq arg\times arg$ and $\bm{acc}\subseteq Pow(arg)$. Given $\bm{att}$, $\bm{acc}$ is a solution to the direct problem if and only if $\bm{acc}$ is an ML prediction in ${\cal M}$ with $w\geq 2$.
\end{theorem}
Theorem \ref{thm:MLandDP} states that an ML prediction in model ${\cal M}$ coincides with a solution to the direct problem. In other words, a direct problem is just one problem we can deal with using the abstract argumentation model.
\par
This fact causes the question, what type of direct problems the abstract argumentation model allows us to deal with in general. The answer is that the abstract argumentation model may discuss a solution to a direct problem depending on multiple attack relations (recall Equation (\ref{eq:af4})). This is because attack relations are assumed to be probabilistically distributed in the abstract argumentation model. As for the \emph{evidence} or marginal likelihood, i.e., $p(\bm{Acc})$, we have
\begin{eqnarray}
p(\bm{Acc})&=&\sum_{att_{1}}\sum_{att_{2}}\cdots\sum_{att_{M}}p(\bm{Acc},att_{1},att_{2},\cdots,att_{M})\nonumber\\
&=&\bm{\sum_{\bm{att}}}p(\bm{Acc},\bm{att})\nonumber\\
&=&\bm{\sum_{\bm{att}}}p(\bm{Acc}|\bm{att})p(\bm{att}).\label{eq:8}
\end{eqnarray}
We now suppose that an attack relation is deterministic, and thus, there is an attack relation $\hat{\bm{att}}$ such that $p(\hat{\bm{att}})=1$ holds. It can be seen as a restriction on the attack distribution $p(\bm{Att})$. We then have
\begin{eqnarray*}
p(\bm{Acc})=p(\bm{Acc}|\hat{\bm{att}})=\langle p(\bm{acc}_{1}|\hat{\bm{att}}),p(\bm{acc}_{2}|\hat{\bm{att}}),\cdots,p(\bm{acc}_{N}|\hat{\bm{att}}) \rangle
\end{eqnarray*}
where $N$ is the number of the acceptability variables. It states that the \emph{likelihood distribution}, i.e., $p(\bm{Acc}|\hat{\bm{att}})$, is a special form of the evidence under the restriction. Now, we pay attention to the acceptability $\bm{\hat{acc}}$ maximizing the likelihood. It can be seen as a restriction on the likelihood distribution. It is represented by
%
%
\begin{eqnarray*}
\bm{\hat{acc}}=\argmax_{\bm{acc}}p(\bm{acc}|\hat{\bm{att}}).
\end{eqnarray*}
This is an ML prediction. It thus states that an ML prediction is a special form of a prediction with a likelihood distribution under the restriction.
\par
The theoretical consequences discussed in this section are summarized as follows. An ML prediction is equivalent to a solution to a direct problem. An ML prediction is a special case of a prediction with a likelihood distribution $p(\bm{Acc}|\bm{att})$. A prediction with a likelihood distribution is a special case of a prediction with an evidence $p(\bm{Acc})$. The practical implications of the consequences are summarized as follows.
\begin{itemize}
\item There is no positive reason to use an ML prediction instead of a deterministic approach to solve a direct problem.
\item It is reasonable to use a likelihood distribution instead of an ML prediction because it tells us the uncertainty to what extent a set of arguments is acceptable.
\item It is reasonable to use an evidence instead of a likelihood distribution because it gives a formal account of acceptability caused by multiple attack relations distributed probabilistically.
\end{itemize}
\par
Figure \ref{fig:illustration} illustrates an advantage of the use of Bayesian approach in the direct problem. It shows that a prediction of acceptability is performed by taking into account all attack relations distributed probabilistically. A deterministic solution to the direct problem is obviously its special case where there is only one attack relation whose posterior probability is one.
%
\section{Empirical Evaluation of Model Correctness}
\subsection{Posterior Acceptability Distribution}
%
Our basic observations are that an attack relation is not observable, and that an acceptability of arguments is observable via a vote in various social networking services. We thus investigate to what extent the posterior distribution over attack relations given an acceptability of arguments successfully predicts another acceptability of arguments.\footnote{One might think that a solution to the inverse problem should be evaluated based on the fact that the solution conforms to a human judgement. As we briefly discussed in Section 1, however, it is not a fundamental requirement of the inverse problem. The inverse problem basically asks for an attack relation explaining an acceptability well in terms of acceptability semantics.} The acceptability of a set $e$ of arguments is predicted using its posterior predictive distribution given as follows.
%
%
%
\begin{eqnarray}
p(Acc_{e}|\bm{acc})&=&\sum_{att_{1}}\sum_{att_{2}}...\sum_{att_{M}}p(Acc_{e},att_{1},att_{2},...,att_{M}|\bm{acc})\nonumber\\
&=&\bm{\sum_{\bm{att}}}p(Acc_{e},\bm{att}|\bm{acc})\nonumber\\
&=&\bm{\sum_{\bm{att}}}p(Acc_{e}|\bm{att})p(\bm{att}|\bm{acc})\nonumber\\
&=&\bm{\sum_{\bm{att}}}\theta_{e|\bm{att}}^{Acc_{e}}(1-\theta_{e|\bm{att}})^{1-Acc_{e}}p(\bm{att}|\bm{acc}).\label{eq:exactprediction}
\end{eqnarray}
In line 3, the left term is the acceptability likelihood given the attack relation and the right term is the posterior probability of the attack relation. It thus can be seen that the posterior probability functions as a weight to its likelihood.
\par
\begin{figure}[t]
\begin{center}
 \includegraphics[scale=0.3]{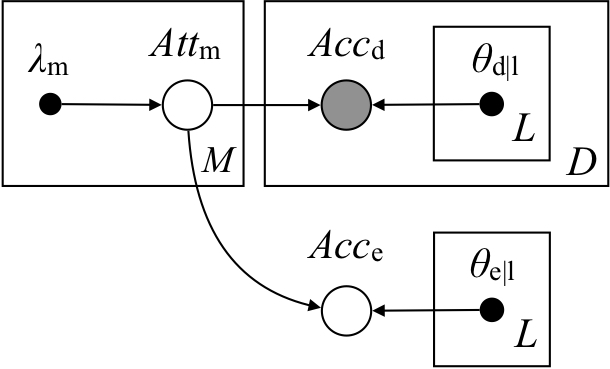}
  \caption{Abstract argumentation model used when predicting an unknown acceptability. In practice, it first traces the causality back to the attack relation from the observed acceptability, i.e., $Att_{m}$ from $Acc_{d}$. It then follows the causality to derive the distribution over unobserved acceptability, i.e., $Acc_{e}$.}
  \label{predictionmodel}
\end{center}
\end{figure}

Figure \ref{predictionmodel} shows the Bayesian network used in the prediction. It is an instance of the model shown in Figure \ref{argumentationmodel}. We assume $\theta_{e|\bm{att}}$ to be the linear acceptability parameter. Recall Definition \ref{def:LAP}. It has the form
%
\begin{eqnarray*}
\theta_{e|\bm{att}}=\max_{d\in\varepsilon(arg,\bm{att})} \left\{ \frac{|tp(e,d)|+|tn(e,d)|}{|arg|}\right\}.
\end{eqnarray*}
We use the linear acceptability parameter because it represents an \emph{accuracy}, a machine learning criterion often used to evaluate predictive performance. Indeed, the inside of the bracket represents an accuracy of $d$ with respect to $e$. Since $d$ is an extension caused by $\bm{att}$, it is reasonable to think that the above equation represents an accuracy of $\bm{att}$ with respect to $e$. Now, the posterior predictive distribution over the acceptability of $e$ is given by
%
%
\begin{eqnarray*}
p(Acc_{e}=1|\bm{acc})&=&\bm{\sum_{\bm{att}}}\max_{d\in\varepsilon(arg,\bm{att})} \left\{ \frac{|tp(e,d)|+|tn(e,d)|}{|arg|}\right\}p(\bm{att}|\bm{acc})\\
p(Acc_{e}=0|\bm{acc})&=&1-p(Acc_{e}=1|\bm{acc}).
\end{eqnarray*}
Therefore, the whole accuracy is the weighted sum of the accuracies of all possible attack relations with respect to $e$. Intuitively speaking, each accuracy has an influence on the whole accuracy as much as the posterior probability of the attack relation. It is a natural extension of the accuracy criterion when attack relations are distributed probabilistically.
%
%

\subsection{Approximate Inference of Attack Relations}
It is practically difficult to solve Equation (\ref{eq:exactprediction}) exactly due to computational complexity. We thus solve 
%
\begin{eqnarray}
p(Acc_{e}|\bm{acc})&\simeq&\bm{\sum_{\bm{att}}}p(Acc_{e}|\bm{att})\hat{p}(\bm{att}|\bm{acc})\nonumber\\
&=&\bm{\sum_{\bm{att}}}\theta_{e|\bm{att}}^{Acc_{e}}(1-\theta_{e|\bm{att}})^{1-Acc_{e}}\hat{p}(\bm{att}|\bm{acc})\label{eq:prediction},
\end{eqnarray}
where $\hat{p}(\bm{att}|\bm{acc})$ represents an approximation of the true posterior $p(\bm{att}|\bm{acc})$ and $\simeq$ denotes an approximation of the equal sign. We obtain the approximate distribution using Gibbs sampling \cite{geman} that is a simple and widely applicable Markov chain Monte Carlo algorithm. A Gibbs sampling  repeatedly updates a value of each random variable one by one using its posterior distribution given values of all remaining random variables. For example, in the $(i+1)$-th iteration of a Gibbs sampling, it samples a value of each attack relation variable as follows.
%
%
%
\begin{eqnarray*}
att_{1}^{i+1}&\sim& p(Att_{1}| att_{2}^{i}, att_{3}^{i}, \cdots, att_{M}^{i},\bm{acc})\\
att_{2}^{i+1}&\sim& p(Att_{2}| att_{1}^{i+1}, att_{3}^{i}, \cdots, att_{M}^{i},\bm{acc})\\
\vdots\\
att_{M}^{i+1}&\sim& p(Att_{M}| att_{1}^{i+1}, att_{2}^{i+1}, \cdots, att_{M-1}^{i+1},\bm{acc})
\end{eqnarray*}
Here, $\sim$ denotes that the left value is sampled from the right distribution. Let $\bm{att}_{\setminus m}^{(i+1)}$ denote all values except $att_{m}$ sampled in the $(i+1)$-th iteration, i.e., $\bm{att}_{\setminus m}^{(i+1)}=(att_{1}^{i+1}, att_{2}^{i+1},\cdots, att_{m-1}^{i+1}, att_{m+1}^{i},\cdots,att_{M}^{i})$. In general, the expression used in an attack relation sampling is given by 
%
%
%
\begin{eqnarray*}
&&p(Att_{m}|\bm{att}_{\setminus m}^{(i+1)},\bm{acc})\\
&=&\frac{p(\bm{acc}|Att_{m},\bm{att}_{\setminus m}^{(i+1)})p(Att_{m})p(\bm{att}_{\setminus m}^{(i+1)})}{p(\bm{att}_{\setminus m}^{(i+1)},\bm{acc})}\\
&\propto&p(\bm{acc}|Att_{m},\bm{att}_{\setminus m}^{(i+1)})p(Att_{m})\\
&=&\lambda_{m}^{Att_{m}}(1-\lambda)^{1-Att_{m}}\prod_{d}^{D}\theta_{d|Att_{m},\bm{att}_{\setminus m}^{(i+1)}}^{acc_{d}}(1-\theta_{d|Att_{m},\bm{att}_{\setminus m}^{(i+1)}})^{1-acc_{d}}.
\end{eqnarray*}
%
%
Here, Bayes' theorem was used in line 2, the denominator irrelevant to $Att_{m}$ was eliminated in line 3, and the parameters of the Bernoulli distributions were introduced in line 4.
\par
The Gibbs sampling algorithm is shown in Algorithm \ref{algorithm}. Lines 3-10 show a generation process of a specific value of every attack relation. This process is iterated $I$ times so that it yields a distribution of histogram, denoted by {\bf\textit{freq}}, approximating the true posterior. The algorithm returns the approximate distribution obtained by normalizing the histogram. 
%
%
\begin{algorithm}[t]
\caption{Gibbs sampling for the abstract argumentation model}\label{algorithm}
{\small
\begin{algorithmic}[1]
\Require{Observation $\bm{acc}$, semantics $\varepsilon$, constant $w$ of the exponential acceptability parameters, constant $\lambda_{m}$ of the attack parameters, iteration number $I$ and burn-in period $B$}
\Ensure{An approximation of the true posterior distribution over attack relations}
%
\State{Get $\bm{att}^{(0)}$ by randomly assigning $0$ or $1$ to all elements of $\bm{Att}$}
\For{$i=0$ to $I$}
	\ForAll{$Att_{m}\in \bm{Att}$}
		\State {\bf\textit{prob}} $\leftarrow[1-\lambda_{m},\lambda_{m}]$ \Comment{Compute $p(att_{m}^{(i+1)})$}
		\ForAll{$acc_{d}\in \bm{acc}$} \Comment{Compute $p(acc_{d}|att_{m}^{(i+1)},\bm{att}_{\setminus m}^{(i+1)})$}
			\State $prob[0] \leftarrow prob[0] \cdot \theta_{d|Att_{m}=0,\bm{att}_{\setminus m}^{(i+1)}}^{acc_{d}}(1-\theta_{d|Att_{m}=0,\bm{att}_{\setminus m}^{(i+1)}})^{1-acc_{d}}$
			\State $prob[1] \leftarrow prob[1] \cdot \theta_{d|Att_{m}=1,\bm{att}_{\setminus m}^{(i+1)}}^{acc_{d}}(1-\theta_{d|Att_{m}=1,\bm{att}_{\setminus m}^{(i+1)}})^{1-acc_{d}}$			
		\EndFor
	\State $att_{m}^{(i+1)}\sim$ {\bf\textit{prob}} \Comment{Generate $att_{m}^{(i+1)}$ from $p(Att_{m}|\bm{att}_{\setminus m}^{(i+1)},\bm{acc})$}
	\EndFor
\EndFor
\State {\bf\textit{freq}} $\leftarrow\emptyset$
\ForAll{$\bm{att}\in\{\bm{att}^{(i)}|B< i\leq I\}$}\Comment{Compute an attack relation histogram}
	\State $count\leftarrow$ the number of occurrence of $\bm{att}$ in $(\bm{att}^{(i)}|B< i\leq I)$
	\State {\bf\textit{freq}} $\leftarrow$ {\bf\textit{freq}} $\cup\{(count,\bm{att})\}$
\EndFor
\State{\textbf{return} ($count/(I-B)| (count,\bm{att})\in$ {\bf\textit{freq}})}
\end{algorithmic}
}
\end{algorithm}

\subsection{Learning Performance Evaluation}
\begin{figure}[t]
\begin{center}
 \includegraphics[scale=0.4]{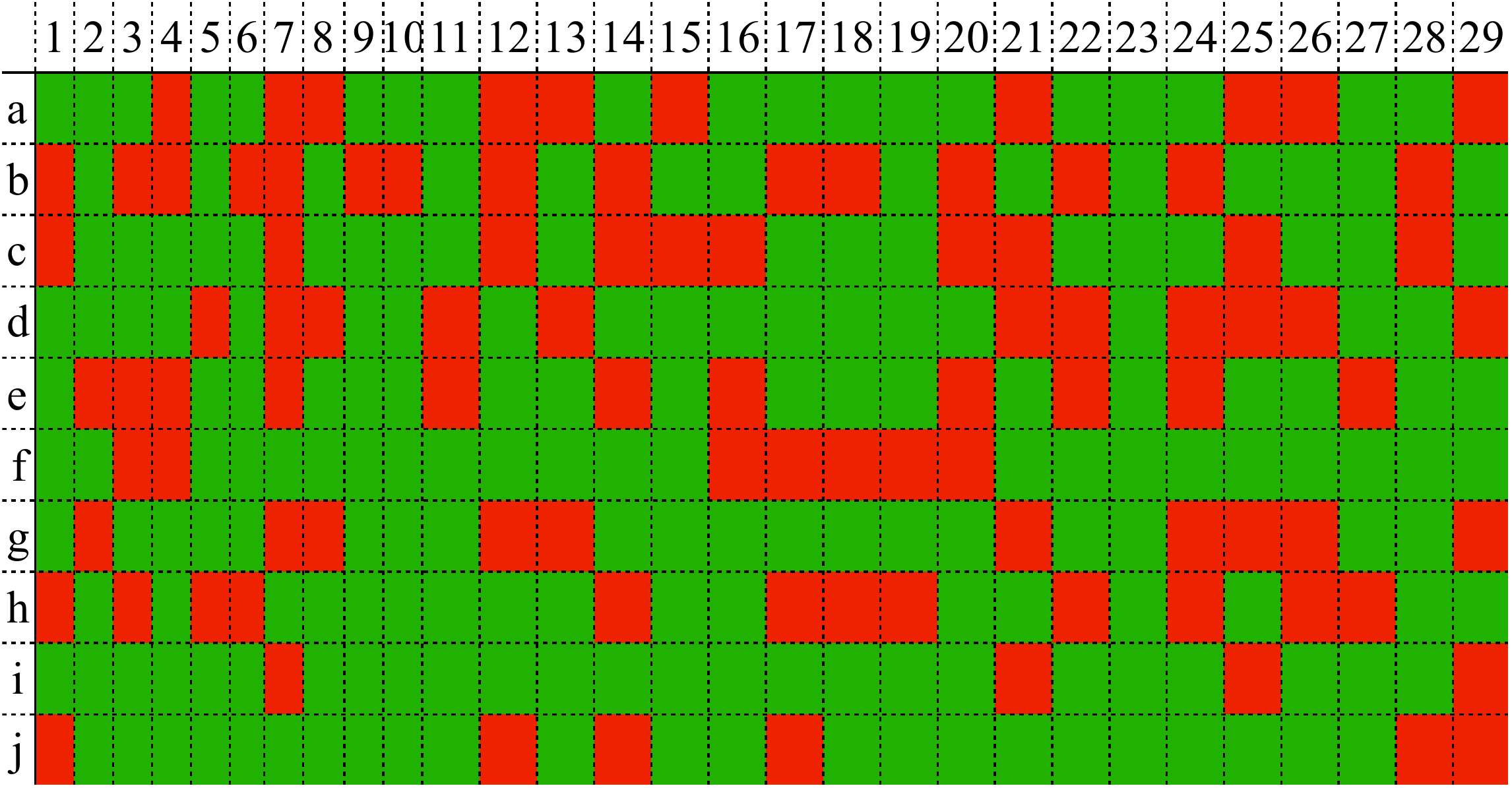}
  \caption{Twenty-nine anonymous participants' sentiments on ten individual arguments. Each green and red cell denotes that the participant agrees and disagrees the argument, respectively.}
  \label{dataset}
\end{center}
\end{figure}
Figure \ref{dataset} shows our dataset consisting of twenty-nine participants' sentiments regarding acceptability of individual ten arguments manually extracted from an online forum. Their textual contents are found in Appendix. Each anonymous participant is presented all of the arguments before she expresses her sentiments. We use the following parameters required in Algorithm \ref{algorithm}: $\varepsilon$ as complete semantics, $w=100$, $\lambda_{\{a,b\}}=0.5$, for all arguments $a$ and $b$, $I=100$ and $B=0$.
\par
A predictive accuracy is evaluated using a \emph{cross validation}. It divides the dataset into training and test sets disjoint each other. Training set $\bm{acc}$ is used to calculate the approximate posterior attack distribution, i.e., $\hat{p}(\bm{Att}|\bm{acc})$ shown in (\ref{eq:prediction}). Each test set $Acc_{e}$ is used to calculate the posterior acceptability distribution, i.e., $p(Acc_{e}|\bm{Att})$ shown in (\ref{eq:prediction}).
\par
\begin{figure}[t]
\begin{center}
 \includegraphics[scale=0.25]{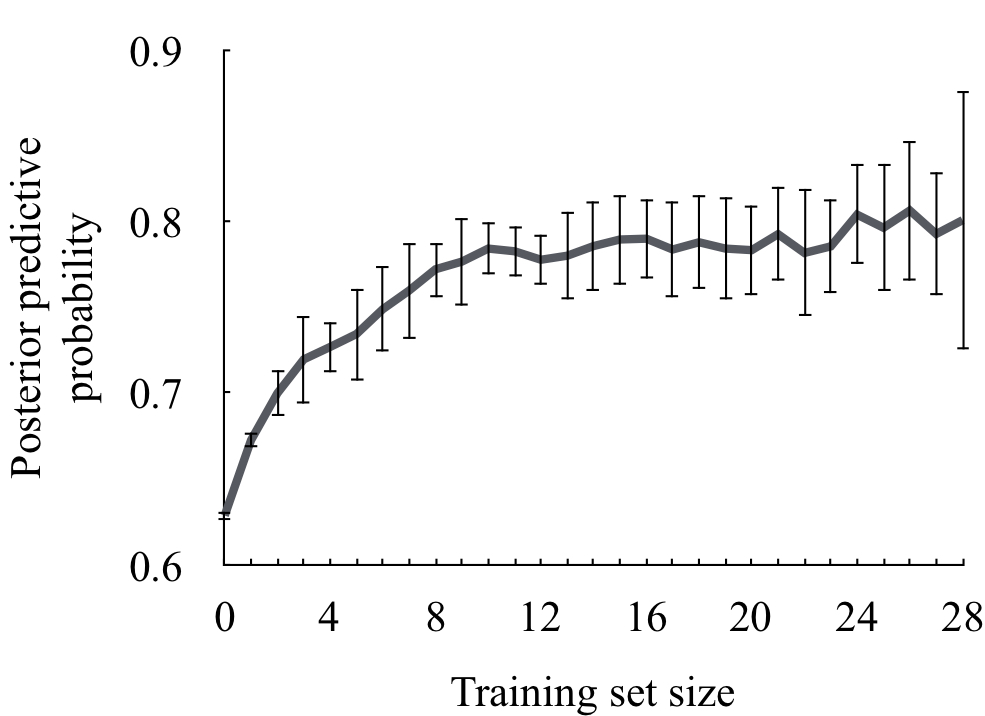}
  \caption{The horizontal axis shows the number of training data used to calculate the posterior attack distribution. The vertical axis shows the posterior acceptability probability, i.e., the predictive accuracy, of test data. For each training set size specified in the horizontal axis, we randomly generated ten divisions of training and test data. The curve is the average of the ten predictive accuracies. Each error bar shows the standard deviation.}
  \label{fig:PPP}
\end{center}
\end{figure}

%
Figure \ref{fig:PPP} shows a learning curve of the dataset obtained with the cross validation. The result shows that the accuracy of the prediction is around $80\%$ at best, on average. It also shows that model ${\cal M}$ predicts the acceptability of test data better when it observes more training data. It implies that participants' sentiments on the acceptability of arguments can be explained fairly using the acceptability semantics.
%
\par
\begin{figure}[t]
\begin{center}
 \includegraphics[scale=0.25]{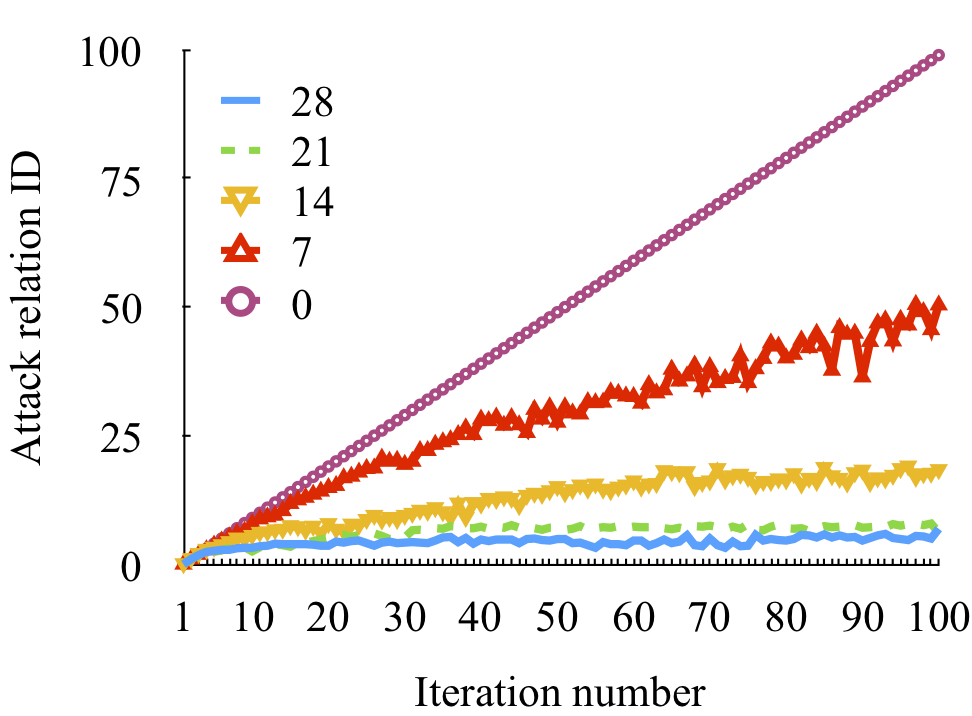}
  \caption{The convergence of the posterior distribution over attack relations. Each series shows the size of training data. The horizontal axis shows the number of iterations in Gibbs sampling and the vertical axis shows the number of occurrences of new attack relations during the sampling.}
  \label{fig:convergence}
\end{center}
\end{figure}

Figure \ref{fig:convergence} shows the convergence of attack relations sampled during the Gibbs sampling. It is observed that sampled attack relations diverge completely when no training data is given. However, they converge when enough training data is provided. It implies that the training data contributes to find an attack relation explaining the data well.
\par
\begin{figure}[t]
\begin{center}
 \includegraphics[scale=0.35]{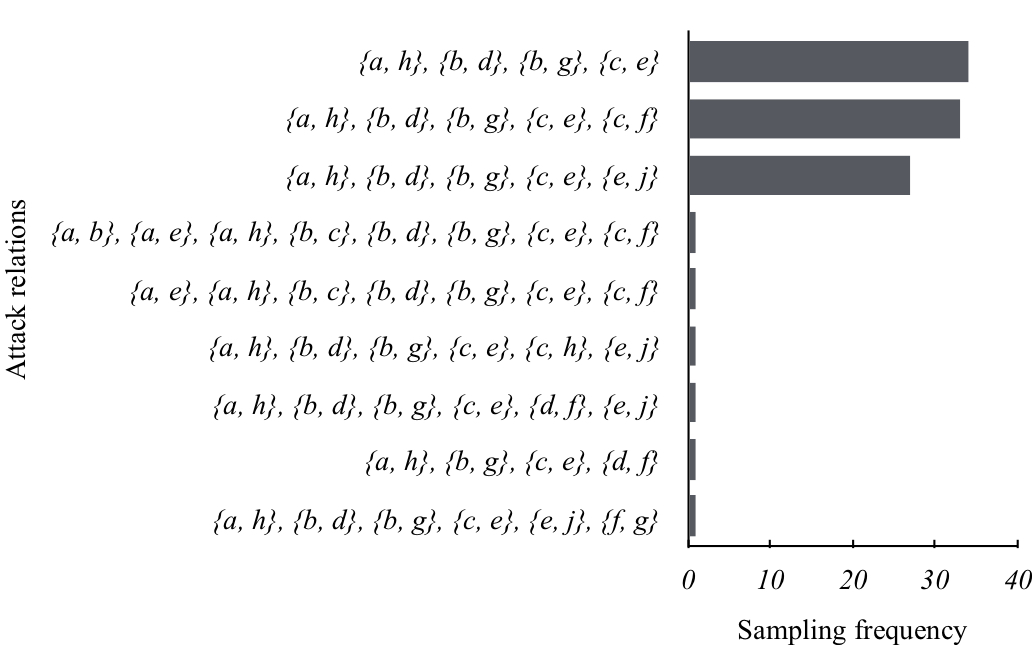}
  \caption{Attack relations sampled versus the number of their occurrences.}
  \label{fig:AFs}
\end{center}
\end{figure}

Figure \ref{fig:AFs} shows all attack relations sampled while one hundred iterations of the Gibbs sampling. Here, we have used all dataset shown in Figure \ref{dataset} as a training set. The textual contents of each argument are shown in Appendix.

\section{Conclusions and Discussion}
In this paper, we have studied an abstract but general probabilistic model of the theory of abstract argumentation \cite{dung:95}. It captures the central notion of acceptability of arguments in a probabilistic way. It allows us to deal with both direct and inverse abstract argumentation problems where attack relations are probabilistically distributed. We showed that a solution to both direct and inverse problems was just a special case of probabilistic inference on the probabilistic model. We empirically demonstrated that our Bayesian solution to an inverse problem was qualified in the sense that it fairly predicts human sentiments regarding acceptability of arguments.
\par
From the perspective of data-driven prediction of an attack relation, one might think that acceptability semantics should not be specified in advance but estimated from data. Such problem is often classified into a \emph{model identification problem} \cite{aster} in contrast to an inverse and direct problem. In the context of the abstract argumentation, it aims to determine acceptability semantics given attack relations and argument acceptability. Although we can think of a problem simultaneously discussing both inverse problem and model identification problem, an outcome we can get via the discussion is not very practical in terms of computational complexity and thus applicability. It moreover contradicts the goal of this paper providing an abstract but general probabilistic model dealing with a direct and inverse problem.
%
%
%
\par
Our simple probabilistic model has important contributions. The past two decades in the field of computational argumentation in AI witnessed an intensive study of acceptability semantics and dialectical proof theories for various interpretations and derivations of a consequence associated with a given argumentation framework, a knowledge representation for argumentative knowledge, e.g., \cite{cayrol,amgoud:09,bench-capon:02,modgil,leite,verheij,caminada,dung:06,baroni:05,coste-marquis}. This paper places them on a direct problem and raises a new research direction, an inverse problem. An inverse problem cannot be solved without the mathematics of a direct problem. This paper thus finds another value of their studies and functions as a guideline on how to make use of their formalisms to deal with their inverse problems. Moreover, this paper lays the foundation for making acceptability semantics data-driven. A data-driven approach is a minimal requirement at the current era of data science. Data science expects AI to gain insight into data available on the web or via sensors in the real world. From data science point of view, a weakness of the study of the abstract argumentation, and a symbolic AI in general, is a knowledge acquisition bottleneck that is a problem on how to acquire knowledge from data. This is because, in general, a knowledge representation is not what one has, but it is what one wants and can have as a result of problem analysis. An input of an inverse problem is a sentiment regarding acceptability of arguments. In contrast to a direct problem, it is an unstructured data and thus available on the web, e.g., votes in various social networking services. It makes us easier to find a killer AI application of acceptability semantics.
%
\section*{Acknowledgements}
The authors are grateful to Martin Caminada for his support, encouragement and valuable discussion. This work was supported by JSPS KAKENHI Grant Number 18K11428.
\bibliographystyle{elsarticle-num}
\bibliography{btxkido}
\section*{Appendix}
\subsection*{Proofs of Theorems}
%
\begin{proof}[Proposition \ref{prop:1}]
This directly follows from the fact that, for any two abstract argumentation frameworks $AF_{1}$ and $AF_{2}$ with the same set of arguments, and symmetric and irreflexive attack relations, if $\varepsilon(AF_{1})=\varepsilon(AF_{2})$ then $AF_{1}=AF_{2}$ for any semantics except grounded semantics. As for grounded semantics, it is obvious because the empty set is the grounded extension of any $AF_{i}$ ($i=1,2$) with a symmetric non-empty attack relation. For the remaining semantics, since $AF_{i}$ is symmetric and irreflexive, it can be regarded as an undirected graph without self-loop where a node and an edge represent an argument and an attack between arguments, respectively. For all independent sets $S$ of this graph, $S$ is an admissible set of $AF_{i}$. There is thus a set $T\supseteq S$ such that $T$ is a preferred extension $AF_{i}$. Since $AF_{i}$ is symmetric, a set of arguments is a preferred extension if and only if it is a stable extension. It is moreover obvious that a preferred extension is a complete extension. If $AF_{1}\neq AF_{2}$ then the set of independent sets of $AF_{1}$ does not coincide with that of $AF_{2}$. Therefore, the set of preferred extensions of $AF_{1}$ does not coincide with that of $AF_{2}$.
\end{proof}
\begin{proof}[Proposition \ref{prop:2}]
Let $f$ be the normalized exponential function, i.e., Equation (\ref{eq:nef}). We then have
\begin{eqnarray*}
\lim_{w\rightarrow\infty}f(x)=\lim_{w\rightarrow\infty}\frac{w^{x-|arg|}-\frac{1}{w^{|arg|}}}{1-\frac{1}{w^{|arg|}}}=\lim_{w\rightarrow\infty}w^{x-|arg|}=
\begin{cases}
1& x=|arg|\\
0& otherwise.
\end{cases}
\end{eqnarray*}
\end{proof}
\begin{proof}[Proposition \ref{prop:3}]
Let $f$ be the normalized exponential function, i.e., Equation (\ref{eq:nef}). Using l'H\r{o}pital's rule, we have
\begin{eqnarray*}
\lim_{w\rightarrow\infty}f(x)=\lim_{w\rightarrow1}\frac{w^{x}-1}{w^{|arg|}-1}=\lim_{w\rightarrow1}\frac{xw^{x-1}}{|arg|w^{|arg|-1}}=\frac{x}{|arg|}.
\end{eqnarray*}
\end{proof}
\begin{proof}[Theorem \ref{thm:IPtoML}]
Let $\bm{att}^{\mathrm{k}}$ be an arbitrary attack relation. $\bm{acc}$ can be divided into four disjoint sets.
\begin{eqnarray*}
\bm{acc}^{tp}&=&\{(Acc_{d}=1)\in\bm{acc}|d\in\varepsilon(arg,\bm{att}^{\mathrm{k}},\bm{att})\}\\
\bm{acc}^{fn}&=&\{(Acc_{d}=1)\in\bm{acc}|d\notin\varepsilon(arg,\bm{att}^{\mathrm{k}},\bm{att})\}\\
\bm{acc}^{fp}&=&\{(Acc_{d}=0)\in\bm{acc}|d\in\varepsilon(arg,\bm{att}^{\mathrm{k}},\bm{att})\}\\
\bm{acc}^{tn}&=&\{(Acc_{d}=0)\in\bm{acc}|d\notin\varepsilon(arg,\bm{att}^{\mathrm{k}},\bm{att})\}
\end{eqnarray*}
Since acceptability is independent and identically distributed, we have
\begin{eqnarray*}
&&p(\bm{acc}|\bm{att}^{\mathrm{k}},\bm{att})\\
&=&p(\bm{acc}^{tp}|\bm{att}^{\mathrm{k}},\bm{att})p(\bm{acc}^{fn}|\bm{att}^{\mathrm{k}},\bm{att})p(\bm{acc}^{fp}|\bm{att}^{\mathrm{k}},\bm{att})p(\bm{acc}^{tn}|\bm{att}^{\mathrm{k}},\bm{att})\\
&=&\prod_{d_{1}}^{|\bm{acc}^{tp}|}\theta_{d_{1}|\bm{att}^{\mathrm{k}},\bm{att}}\prod_{d_{2}}^{|\bm{acc}^{fn}|}\theta_{d_{2}|\bm{att}^{\mathrm{k}},\bm{att}}\prod_{d_{3}}^{|\bm{acc}^{fp}|}(1-\theta_{d_{3}|\bm{att}^{\mathrm{k}},\bm{att}})\prod_{d_{4}}^{|\bm{acc}^{tn}|}(1-\theta_{d_{4}|\bm{att}^{\mathrm{k}},\bm{att}}).
\end{eqnarray*}
%
Since $\bm{att}$ is a solution to the inverse problem, i.e., $\bm{acc}=\varepsilon(arg,\bm{att}^{\mathrm{k}},\bm{att})$, $\bm{acc}^{tp}\cup\bm{acc}^{tn}=\bm{acc}$ and $\bm{acc}^{fp}\cup\bm{acc}^{fn}=\emptyset$ holds. We thus have 
\begin{eqnarray*}
p(\bm{acc}|\bm{att}^{\mathrm{k}},\bm{att})&=&\prod_{d_{1}}^{|\bm{acc}^{tp}|}\theta_{d_{1}|\bm{att}^{\mathrm{k}},\bm{att}}\prod_{d_{4}}^{|\bm{acc}^{tn}|}(1-\theta_{d_{4}|\bm{att}^{\mathrm{k}},\bm{att}}).
\end{eqnarray*}
Now, any attack relation that is not a solution to the inverse problem causes a shift of an element from $\bm{acc}^{tp}$ to $\bm{acc}^{fn}$ or $\bm{acc}^{tn}$ to $\bm{acc}^{fp}$. However, this never makes the probability higher. This is because $\theta_{d_{1}|\bm{att}^{\mathrm{k}},\bm{att}}>\theta_{d_{2}|\bm{att}^{\mathrm{k}},\bm{att}}$ and $(1-\theta_{d_{4}|\bm{att}^{\mathrm{k}},\bm{att}})>(1-\theta_{d_{3}|\bm{att}^{\mathrm{k}},\bm{att}})$ hold from Definition \ref{def:eap} where $\theta_{d_{1}|\bm{att}^{\mathrm{k}},\bm{att}}=1$, $\theta_{d_{2}|\bm{att}^{\mathrm{k}},\bm{att}}< 1$, $\theta_{d_{3}|\bm{att}^{\mathrm{k}},\bm{att}}=1$ and $\theta_{d_{4}|\bm{att}^{\mathrm{k}},\bm{att}}< 1$ hold.
\end{proof}
\begin{proof}[Theorem \ref{thm:MLtoIP}]
It is enough to show a counterexample. Given $\bm{acc}=\{\emptyset,\{a,b\}\}$ and $\bm{att}^{\mathrm{k}}=\emptyset$, there is an ML estimation $att_{(a,b)}$ because $p(\bm{acc}|\bm{att}^{\mathrm{k}},att_{(a,b)})\geq 0$ holds for any $att_{(a,b)}$. However, it is never a solution to the inverse problem because there is no attack relation of which $\bm{acc}$ is the set of extensions.
\end{proof}
\begin{proof}[Theorem \ref{thm:MLandDP}]
($\Rightarrow$) $\bm{acc}$ can be divided into four disjoint sets.
\begin{eqnarray*}
\bm{acc}^{tp}&=&\{(Acc_{d}=1)\in\bm{acc}|d\in\varepsilon(arg,att)\}\\
\bm{acc}^{fn}&=&\{(Acc_{d}=1)\in\bm{acc}|d\notin\varepsilon(arg,att)\}\\
\bm{acc}^{fp}&=&\{(Acc_{d}=0)\in\bm{acc}|d\in\varepsilon(arg,att)\}\\
\bm{acc}^{tn}&=&\{(Acc_{d}=0)\in\bm{acc}|d\notin\varepsilon(arg,att)\}
\end{eqnarray*}
Since acceptability is independent and identically distributed, we have
\begin{eqnarray*}
p(\bm{acc}|\bm{att})&=&p(\bm{acc}^{tp}|\bm{att})p(\bm{acc}^{fn}|\bm{att})p(\bm{acc}^{fp}|\bm{att})p(\bm{acc}^{tn}|\bm{att})\\
&=&\prod_{d_{1}}^{|\bm{acc}^{tp}|}\theta_{d_{1}|\bm{att}}\prod_{d_{2}}^{|\bm{acc}^{fn}|}\theta_{d_{2}|\bm{att}}\prod_{d_{3}}^{|\bm{acc}^{fp}|}(1-\theta_{d_{3}|\bm{att}})\prod_{d_{4}}^{|\bm{acc}^{tn}|}(1-\theta_{d_{4}|\bm{att}}).
\end{eqnarray*}
Since $\bm{att}$ is a solution to the direct problem, $\bm{acc}=\varepsilon(arg,\bm{att})$ holds. Thus, $\bm{acc}^{tp}\cup\bm{acc}^{tn}=\bm{acc}$ and $\bm{acc}^{fp}\cup\bm{acc}^{fn}=\emptyset$ holds. We thus have
\begin{eqnarray}
p(\bm{acc}|\bm{att})&=&\prod_{d_{1}}^{|\bm{acc}^{tp}|}\theta_{d_{1}|\bm{att}}\prod_{d_{4}}^{|\bm{acc}^{tn}|}(1-\theta_{d_{4}|\bm{att}})\label{eq:MLandDP}
\end{eqnarray}
Now, any acceptability that is not a solution to the direct problem causes a shift of an element from $\bm{acc}^{tp}$ to $\bm{acc}^{fp}$ or $\bm{acc}^{tn}$ to $\bm{acc}^{fn}$. However, this never makes the probability higher because $\theta_{d_{1}|\bm{att}}>(1-\theta_{d_{3}|\bm{att}})$ and $(1-\theta_{d_{4}|\bm{att}})>\theta_{d_{2}|\bm{att}}$ hold. This is because, from Definition \ref{def:eap}, if $w\geq 2$ holds then $\theta_{d_{1}|\bm{att}}=1$, $\theta_{d_{2}|\bm{att}}<0.5$, $\theta_{d_{3}|\bm{att}}=1$ and $\theta_{d_{4}|\bm{att}}<0.5$ hold. Here, we prove $\theta_{d|\bm{att}}< 0.5$ holds if $d\notin\varepsilon(arg,\bm{att})$ and $w\geq 2$, as follows. Let $d$ be a set of arguments such that there is, at best, an extension $e\in\varepsilon(arg,att)$ satisfying $|tp(d,e)|+|tn(d,e)|=|arg|-1$. Here, we do not need to think of $|tp(d,e)|+|tn(d,e)|<|arg|-1$ because of the monotonicity of $\theta_{d|\bm{att}}$. We then have
\begin{eqnarray*}
\theta_{d|\bm{att}}&=&\frac{w^{|arg|-1}-1}{w^{|arg|}-1}=\frac{1}{w}\frac{w^{|arg|}-w}{w^{|arg|}-1}\\
2\theta_{d|\bm{att}}&=&\frac{2}{w}\frac{w^{|arg|}-w}{w^{|arg|}-1}.
\end{eqnarray*}
Now, $2\theta_{d|\bm{att}}< 1$ holds, for all $w\geq 2$. Indeed, if $w=2$ then $2/w=1$ and $(w^{|arg|}-w)/(w^{|arg|}-1)<1$ hold. If $w>2$ then $2/w<1$ and $(w^{|arg|}-w)/(w^{|arg|}-1)<1$ hold as well.
\par
($\Leftarrow$) We show that if $\bm{acc}$ is not a solution to the direct problem then it does not maximize the likelihood of $\bm{acc}$. We do not need to consider the case where there is no solution to a direct problem because it does not satisfy the antecedent. If $\bm{acc}$ is not a solution then $\bm{acc}^{tp}\cup\bm{acc}^{tn}\subset\bm{acc}$ and $\bm{acc}^{fn}\cup\bm{acc}^{fp}\supset\emptyset$ hold. However, since a direct problem satisfies the solution uniqueness, there is a unique $\bm{acc'}$ such that $\bm{acc'}^{tp}\cup\bm{acc'}^{tn}=\bm{acc'}$ and $\bm{acc'}^{fn}\cup\bm{acc'}^{fp}=\emptyset$. Since $\theta_{d_{1}|\bm{att}}>(1-\theta_{d_{3}|\bm{att}})$ and $(1-\theta_{d_{4}|\bm{att}})>\theta_{d_{2}|\bm{att}}$ hold in Equation (\ref{eq:MLandDP}), $p(\bm{acc'}|\bm{att})>p(\bm{acc}|\bm{att})$ holds.
\end{proof}

\subsection*{Dataset of Arguments}
The ten arguments used in our empirical analysis have the following textual contents. They were presented in this order in SYNCLON \cite{synclon}. We manually extracted them and translated them into English.
%
\begin{description}
\item[a:] \textit{Euthanasia (painless death) should be allowed by law because a medical treatment by doctors should respect patient's will.}
\item[b:] \textit{Euthanasia should not be allowed by law. You assume that one who applies euthanasia is a doctor. I doubt doctor's right to commit a murder.}
\item[c:] \textit{I agree with you that euthanasia should be allowed by law, but disagree with the point that one who applies is a doctor. A doctor should always consider the way to cure diseases.}
\item[d:] \textit{If doctor's role is only to cure a disease then they can do nothing for patients with an untreatable disease. I think that a medical treatment should consider death more seriously. Only doctors can apply euthanasia appropriately because they can assess patients' physical and mental state accurately.}
\item[e:] \textit{But, it will be scary if there are professionals for euthanasia.}
\item[f:] \textit{I mean that I disagree with the point that a doctor encourages a patient to choose euthanasia. A doctor can help a patient with an untreatable disease without encouraging her to choose euthanasia. A doctor and a patient can lay heads together to think about how she can live with a disease. This is how a doctor can consider patient's death.}
\item[g:] \textit{Of course, no one has a right to encourage a patient to choose euthanasia. Euthanasia can be applied on the basis of patient's and her family's agreement. I think it is possible that a doctor applies euthanasia when their will for euthanasia is confirmed.}
\item[h:] \textit{I think that an agreement or confirmation does not show patient's true will because I occasionally wish for death when I have a hard experience. But, it sometimes comes from a temporary emotion. I certainly know that my experience is a little thing compared to patient's sufferance. But, I think it is insufficient to apply euthanasia based on patient's will for death.}
\item[i:] \textit{Can you accept legal euthanasia if it is based not only on patient's will, but also her family's will?}
\item[j:] \textit{I cannot accept legal euthanasia even though it is based not only on patient's will, but also her family's will. At the moment, I cannot agree euthanasia without considering patient's physical condition. So, I agree with passive euthanasia.}
\end{description}

\end{sloppypar}
\end{document}